\documentclass{article}

\usepackage[verbose=true,letterpaper]{geometry}
\usepackage{fancyhdr}

\makeatletter

\AtBeginDocument{
  \newgeometry{
    textheight=9in,
    textwidth=6.5in,
    top=1in,
    headheight=14pt,
    headsep=25pt,
    footskip=30pt
  }
}

\fancyheadoffset{0pt}
\def\keywordname{{\bfseries \emph Keywords}}%
\def\keywords#1{\par\addvspace\medskipamount{\rightskip=0pt plus1cm
\def\and{\ifhmode\unskip\nobreak\fi\ $\cdot$
}\noindent\keywordname\enspace\ignorespaces#1\par}}

\renewcommand{\normalsize}{%
  \@setfontsize\normalsize\@xpt\@xipt
  \abovedisplayskip      7\p@ \@plus 2\p@ \@minus 5\p@
  \abovedisplayshortskip \z@ \@plus 3\p@
  \belowdisplayskip      \abovedisplayskip
  \belowdisplayshortskip 4\p@ \@plus 3\p@ \@minus 3\p@
}
\normalsize
\renewcommand{\small}{%
  \@setfontsize\small\@ixpt\@xpt
  \abovedisplayskip      6\p@ \@plus 1.5\p@ \@minus 4\p@
  \abovedisplayshortskip \z@  \@plus 2\p@
  \belowdisplayskip      \abovedisplayskip
  \belowdisplayshortskip 3\p@ \@plus 2\p@   \@minus 2\p@
}
\renewcommand{\footnotesize}{\@setfontsize\footnotesize\@ixpt\@xpt}
\renewcommand{\scriptsize}{\@setfontsize\scriptsize\@viipt\@viiipt}
\renewcommand{\tiny}{\@setfontsize\tiny\@vipt\@viipt}
\renewcommand{\large}{\@setfontsize\large\@xiipt{14}}
\renewcommand{\Large}{\@setfontsize\Large\@xivpt{16}}
\renewcommand{\LARGE}{\@setfontsize\LARGE\@xviipt{20}}
\renewcommand{\huge}{\@setfontsize\huge\@xxpt{23}}
\renewcommand{\Huge}{\@setfontsize\Huge\@xxvpt{28}}

\providecommand{\section}{}
\renewcommand{\section}{%
  \@startsection{section}{1}{\z@}%
                {-2.0ex \@plus -0.5ex \@minus -0.2ex}%
                { 1.5ex \@plus  0.3ex \@minus  0.2ex}%
                {\large\bf\raggedright}%
}
\providecommand{\subsection}{}
\renewcommand{\subsection}{%
  \@startsection{subsection}{2}{\z@}%
                {-1.8ex \@plus -0.5ex \@minus -0.2ex}%
                { 0.8ex \@plus  0.2ex}%
                {\normalsize\bf\raggedright}%
}
\providecommand{\subsubsection}{}
\renewcommand{\subsubsection}{%
  \@startsection{subsubsection}{3}{\z@}%
                {-1.5ex \@plus -0.5ex \@minus -0.2ex}%
                { 0.5ex \@plus  0.2ex}%
                {\normalsize\bf\raggedright}%
}
\providecommand{\paragraph}{}
\renewcommand{\paragraph}{%
  \@startsection{paragraph}{4}{\z@}%
                {1.5ex \@plus 0.5ex \@minus 0.2ex}%
                {-1em}%
                {\normalsize\bf}%
}
\providecommand{\subparagraph}{}
\renewcommand{\subparagraph}{%
  \@startsection{subparagraph}{5}{\z@}%
                {1.5ex \@plus 0.5ex \@minus 0.2ex}%
                {-1em}%
                {\normalsize\bf}%
}

\newlength{\@abovecaptionskip}
\newlength{\@belowcaptionskip}

\renewenvironment{table}
  {\setlength{\abovecaptionskip}{\@belowcaptionskip}%
   \setlength{\belowcaptionskip}{\@abovecaptionskip}%
   \@float{table}}
  {\end@float}

\renewcommand{\footnoterule}{\kern-3\p@ \hrule width 12pc \kern 2.6\p@}
\def\@listi  {\leftmargin\leftmargini}
\def\@listii {\leftmargin\leftmarginii
              \labelwidth\leftmarginii
              \advance\labelwidth-\labelsep
              \topsep  2\p@ \@plus 1\p@    \@minus 0.5\p@
              \parsep  1\p@ \@plus 0.5\p@ \@minus 0.5\p@
              \itemsep \parsep}
\def\@listiii{\leftmargin\leftmarginiii
              \labelwidth\leftmarginiii
              \advance\labelwidth-\labelsep
              \topsep    1\p@ \@plus 0.5\p@ \@minus 0.5\p@
              \parsep    \z@
              \partopsep 0.5\p@ \@plus 0\p@ \@minus 0.5\p@
              \itemsep \topsep}
\def\@listiv {\leftmargin\leftmarginiv
              \labelwidth\leftmarginiv
              \advance\labelwidth-\labelsep}
\def\@listv  {\leftmargin\leftmarginv
              \labelwidth\leftmarginv
              \advance\labelwidth-\labelsep}
\def\@listvi {\leftmargin\leftmarginvi
              \labelwidth\leftmarginvi
              \advance\labelwidth-\labelsep}

\providecommand{\maketitle}{}
\renewcommand{\maketitle}{%
  \par
  \begingroup
    \renewcommand{\thefootnote}{\fnsymbol{footnote}}
    \renewcommand{\@makefnmark}{\hbox to \z@{$^{\@thefnmark}$\hss}}
    \long\def\@makefntext##1{%
      \parindent 1em\noindent
      \hbox to 1.8em{\hss $\m@th ^{\@thefnmark}$}##1
    }
    \thispagestyle{empty}
    \@maketitle
    \@thanks
  \endgroup
  \let\maketitle\relax
  \let\thanks\relax
}

\newcommand{\@toptitlebar}{
  \hrule height 2\p@
  \vskip 0.25in
  \vskip -\parskip%
}
\newcommand{\@bottomtitlebar}{
  \vskip 0.29in
  \vskip -\parskip
  \hrule height 2\p@
  \vskip 0.09in%
}

\providecommand{\@maketitle}{}
\renewcommand{\@maketitle}{%
  \vbox{%
    \hsize\textwidth
    \linewidth\hsize
    \vskip 0.1in
    \@toptitlebar
    \centering
    {\LARGE\sc \@title\par}
    \@bottomtitlebar
    \textsc{}\\
    \vskip 0.1in
    \def\And{%
      \end{tabular}\hfil\linebreak[0]\hfil%
      \begin{tabular}[t]{c}\bf\rule{\z@}{24\p@}\ignorespaces%
    }
    \def\AND{%
      \end{tabular}\hfil\linebreak[4]\hfil%
      \begin{tabular}[t]{c}\bf\rule{\z@}{24\p@}\ignorespaces%
    }
    \begin{tabular}[t]{c}\bf\rule{\z@}{24\p@}\@author\end{tabular}%
  \vskip 0.4in \@minus 0.1in \center{\today}   \vskip 0.2in
  }
}

\renewenvironment{abstract}
{
  \centerline
  {\large \bfseries \scshape Abstract}
  \begin{quote}
}
{
  \end{quote}
}

\makeatother

\usepackage[utf8]{inputenc}
\usepackage[T1]{fontenc}
\usepackage{microtype}
\usepackage{url}
\usepackage{amsmath,amssymb}
\usepackage{graphicx}
\usepackage{booktabs}
\usepackage{tabularx}
\usepackage{multirow}
\usepackage{array}
\usepackage{enumitem}
\usepackage{float}
\usepackage[ruled,vlined]{algorithm2e}
\SetCommentSty{emph}
\SetKwComment{tcc}{$\triangleright$\ }{}
\SetKwFor{ForEach}{for each}{do}{end}
\usepackage[round,authoryear]{natbib}
\usepackage[colorlinks=true,linkcolor=blue,citecolor=blue,urlcolor=blue]{hyperref}
\usepackage{caption}
\usepackage{subcaption}
\usepackage{placeins}

\graphicspath{{figures/}}
\DeclareGraphicsExtensions{.pdf,.png,.jpg,.jpeg}


\newcolumntype{Y}{>{\centering\arraybackslash}X}
\newcommand{\dw}{d_w}
\newcommand{\Cp}{C_p}
\newcommand{\WSS}{\mathrm{WSS}}

\title{Distance-Aware Attention and Wall-Distance\\
Expert Routing for Transformer-Based 3D Flow Prediction}

\author{
 Sanghyeon Kim \\
  Division of Future Vehicle\\
  Korea Advanced Institute of Science and Technology\\
  Daejeon, 34051, Republic of Korea \\
  \texttt{csmd2012@kaist.ac.kr} \\
  \And
 Sunwoong Yang \\
  Department of Mechanical Engineering\\
  Hanyang University\\
  Ansan, 15588, Republic of Korea \\
  \texttt{sunwoongy@hanyang.ac.kr} \\
  \And
 Namwoo Kang \\
  Cho Chun Shik Graduate School of Mobility\\
  Korea Advanced Institute of Science and Technology\\
  Narnia Labs\\
  Daejeon, 34051, Republic of Korea \\
  \texttt{nwkang@kaist.ac.kr} \\
}

\date{September 7, 2026}

\begin{document}
\maketitle

\begin{abstract}
Transformer surrogates for three-dimensional flow prediction
compress an industrial mesh into a small set of tokens from which
every prediction point reads. Two operations then follow, the
retrieval step in which a point gathers information from the
compressed representation and the feed-forward layer that transforms
what it retrieved, and in current backbones both are blind to where
the point sits in the flow. We condition both on wall-related
physical signals. Distance-aware cross-attention (DA-CA) reshapes
each volume query by its wall distance before retrieval, so that a
point deep in the boundary layer draws different geometric
information than one in the outer flow. Surface-volume
mixture-of-experts (SVMoE) replaces the shared feed-forward layer
with a small set of experts, routed by wall distance for volume
points and by local geometry for surface points. Neither mechanism
refers to a construct specific to one architecture, so we apply both
unchanged to AB-UPT and Transolver-3. On DrivAerML with 50 training
cases, a data budget typical of high-fidelity CFD, DA-CA reduces the
volume pressure error by 10.1\%, and DA-CA and SVMoE together reduce
it by 12.5\%; a
wall-distance decomposition shows that DA-CA improves the near-wall
region at some cost in the far region, which SVMoE recovers, and the
volume experts settle into near-wall, transition, and free-stream
bands without routing supervision. To test whether these gains are
tied to the small data budget or to the backbone they were developed
on, we retrain both backbones on 300 cases: the conditioning
improves every field quantity, reducing volume pressure and velocity
errors by 33.1\% and 18.6\% on AB-UPT and by 21.4\% and 21.3\% on
Transolver-3. Because routing is indexed by wall distance rather
than by body type, the design also transfers: under
Leave-One-Body-Out evaluation on DrivAerNet++ it reduces the volume
pressure error on unseen body types by up to 14.2\%.
\end{abstract}

\section{Introduction}
\label{sec:introduction}

For engineering design, predicting a three-dimensional flow field is
not a simple problem of reconstructing hundreds of millions of
solution values. It is a problem of preserving the localized flow
structures that determine performance. Drag, lift, heat transfer, and
pressure loss are controlled by thin boundary layers, separated shear
layers, recirculation zones, and wakes. These regions may occupy only
a small fraction of the computational domain but dominate the
quantities of interest, whereas the surrounding free stream is
typically smooth and substantially easier to approximate. This severe
imbalance creates a fundamental challenge: a model can be accurate
over most of the domain but still fail precisely where critical
engineering decisions are made.

Obtaining such fields at industrial fidelity remains computationally
expensive. In the DrivAerML dataset \citep{Ashton2024DrivAerML}, for
example, each vehicle variant is resolved on roughly 8.8 million
surface cells and 160 million volume cells, and a single hybrid
RANS-LES case runs for about 40 hours on 1536 CPU cores, some 60,000
core-hours, to produce one time-averaged flow field. While this cost
may be acceptable for
the final verification stage of a design cycle, where a small number
of mature candidates are evaluated, it is prohibitive for early-stage
design exploration, where hundreds or thousands of candidate
geometries may need to be screened.

Neural surrogates address this gap by amortizing the cost
of high-fidelity simulation. Once trained on a set of pre-computed
solutions, graph neural operators, neural fields, and
transformer-based models infer new flow fields orders of magnitude
faster than the underlying solver
\citep{Karniadakis2021PIML,Wang2023ScientificML,
Ranade2025DoMINO,Alkin2025ABUPT}. Fast inference alone, however, does not make a
surrogate useful; accuracy in the regions that matter also depends
on whether geometry merely enters as an input or directly shapes the
computation used to produce each prediction. Two decisions are
central: which parts of the geometry should influence each prediction
point, and whether all points should be transformed by the same
mapping or by mappings adapted to their physical regimes.

Both decisions are shaped by how these models achieve scalability. An
industrial mesh cannot be attended over directly, so a transformer
surrogate first compresses the problem into a small set of tokens and
then lets every prediction point read from that compact set. AB-UPT
selects a subset of mesh points as anchor tokens and lets the
remaining points cross-attend to them~\citep{Alkin2025ABUPT}, while
Transolver assigns points to a small number of learned physical-state
slices and recovers each point's features from those
slices~\citep{Wu2024Transolver}. In both, a \emph{retrieval} step
decides which part of the compressed representation influences a
given point, and a \emph{transformation} step, the feed-forward layer
that follows, decides how the retrieved information is mapped.
Because these two sites are properties of the family rather than of
one model, we develop the two mechanisms on AB-UPT, where their
effect on attention is easiest to inspect, and then apply them
unchanged to Transolver-3~\citep{Zhou2026Transolver3}, the most recent member
of the Transolver family, which we adopt as a second baseline.

Scalability, however, does not settle either of the two decisions
above. The retrieval step decides which part of the compressed
representation influences a given point from learned feature
similarity alone, without being told where in the flow that point
sits, that is, how deep it lies in the boundary layer.
The feed-forward layer that follows applies
one mapping to every token: surface and volume tokens share the same
weights, as do volume points in the boundary layer and in the free
stream, despite their different physical balances
\citep{Schlichting2017BoundaryLayer}. Prior surrogates may encode
geometry in token features \citep{Alkin2024UPT,Alkin2025ABUPT} or
inject it into attention
\citep{Adams2025GeoTransolver,Zhdanov2025Erwin}, but geometry can be
present in the representation without directly determining where
information is gathered or which mapping is applied.
Distance-dependent information transfer and regime-dependent
transformation must then be inferred implicitly from the training
data, making them more likely to remain tied to the geometries seen
during training than to physical variables that carry over to new
ones.

This reliance on implicit learning is especially problematic in
high-fidelity CFD, where the cost of supervision limits both the
number of geometries available and the compute that can be spent on
training. Under such budgets, relying on
the network to infer spatial locality and wall-dependent behavior
from examples alone is inefficient. Empirically, in the unconditioned
anchor-based baseline, cross-attention varies only weakly with
distance even though the prediction error stays concentrated in the
near-wall and separated-flow regions
(Section~\ref{sec:daca-motivation}). Moreover, enlarging the training
set beyond roughly 50 cases yields only marginal gains under a fixed
compute budget (Section~\ref{sec:data-scaling}). This does not mean
that
more data cannot help; it means the data at hand is used more
effectively when the relevant physical structure enters the
computation explicitly. Reducing the high-fidelity requirement
itself, by transferring from cheaper low-fidelity solutions or by
adding PDE residuals to the
loss~\citep{Yang2024DDPINN,Yang2025MFDeepONet}, is a complementary
route that we do not pursue here. Together, these observations
motivate direct intervention at the two computational sites
identified above.

We address the two decisions with physical signals matched to their
respective roles. The first decision is what a point gathers from the
surface. A point buried in the boundary layer and a point in the
outer flow do not need the same thing: near the wall the local
orientation of the surface and the shear it carries dominate the
solution, while further out the overall shape matters more, and how
deep a point sits is what separates the two cases. Distance-Aware
Cross-Attention (DA-CA) supplies exactly that, letting each query's
wall distance $\dw$ reshape the query before the attention score is
formed. The second decision is how the gathered information is
transformed. A point inside the boundary
layer and a point in the free stream are governed by different
physics and should not be transformed by the same nonlinear mapping.
Surface-Volume Mixture-of-Experts (SVMoE) therefore replaces the
single shared mapping with branch-specific expert mixtures, routing
volume tokens by their wall distance $\dw$ and surface tokens, for
which $\dw$ is trivially zero, by a local geometric descriptor. The
two mechanisms read the same coordinate in the volume but inject it
at different points: DA-CA changes what a token retrieves, whereas
SVMoE changes how what it retrieved is transformed.

The main contributions are as follows.

\begin{enumerate}[leftmargin=1.6em,label=\arabic*.]
\item \textbf{Distance-Aware Cross-Attention (DA-CA).} In
three-dimensional flow fields, prediction error concentrates in the
thin band of volume points closest to the body, yet conventional
attention scores every query--anchor pair by learned feature
similarity alone, with no notion of where the query sits relative to
the wall. The proposed DA-CA conditions each volume query on its own
wall distance before the score is formed, so that a point deep in the boundary
layer looks for different information than one in the outer flow. On the DrivAerML dataset, this reduces the volume
pressure error by 10.1\%.

\item \textbf{Surface-Volume Mixture-of-Experts (SVMoE).} Near-wall
shear, wake recirculation, and outer flow are governed by different
physics, yet in conventional methods one and the same feed-forward
transformation is applied to all of them. Where DA-CA conditions
what a point gathers, the proposed SVMoE conditions how it transforms
what it gathered. It replaces the single feed-forward network with
a small set of experts and lets each volume point choose among them
by its wall distance, with a separate mixture for surface points.
Without supervision on the assignment, the volume experts organize
into near-wall, transition, and free-stream bands. Combined with
DA-CA, this reduces the volume pressure error by 12.5\%.

\item \textbf{Robustness to data scale and backbone.} The gains
above are measured at the data budget the method targets and on the
backbone it was developed on, which leaves open whether they are
tied to either. Since neither mechanism refers to anchors, slices,
or any other construct specific to one architecture, we retrain both
AB-UPT and Transolver-3, with and without the components, on 300
DrivAerML cases under an identical protocol. The conditioning
improves every field quantity on both backbones, reducing the volume
pressure error by 33.1\% on AB-UPT and 21.4\% on Transolver-3: the
benefit adds to, rather than substitutes for, additional data, and
does not depend on whether the compressed representation is a set
of anchor tokens or of physical-state slices.

\item \textbf{Generalization performance.} Both components are
developed on DrivAerML, so we test whether they carry over to
geometries they were not tuned on. On the DrivAerNet++
dataset, with one car body type held out from training at a time,
the volume pressure error on the unseen body type falls by up to
14.2\%.
\end{enumerate}

\section{Related Work}
\label{sec:related}

\paragraph{Scalable neural surrogates for 3D flow fields.}
Recent surrogates differ mainly
in where geometry enters the computation. UPT compresses irregular
grid or particle inputs into a fixed-size latent representation and
decodes the solution at arbitrary query
locations~\citep{Alkin2024UPT}. AB-UPT extends this framework with
separate geometry, surface, and volume branches, restricting
quadratic self-attention to a compact set of anchor tokens while
non-anchor points retrieve information from the anchors through
cross-attention~\citep{Alkin2025ABUPT}. Transolver maps discretized
points into learned physical-state tokens and attends in that state
space~\citep{Wu2024Transolver}; Transolver++ and Transolver-3 extend
the formulation toward million- and industrial-scale
meshes~\citep{Luo2025TransolverPP,Zhou2026Transolver3}. Others let
geometry shape the computation more directly: Erwin builds
hierarchical spatial neighborhoods with a ball-tree partition and
performs local attention together with cross-neighborhood
interaction~\citep{Zhdanov2025Erwin}, GeoTransolver augments
physical-state attention with multi-scale geometric and
boundary-condition context~\citep{Adams2025GeoTransolver}, and GAOT
couples geometry-aware operator encoders with a transformer
processor~\citep{Wen2025GAOT}. Point- and field-based designs take a
different route: DoMINO builds local and global geometry
representations from coordinates, signed distance, and surface
normals, with separate aggregation networks for surface and volume
quantities~\citep{Ranade2025DoMINO}; Geom-DeepONet adds the signed
distance to the trunk inputs of a deep operator
network~\citep{He2024GeomDeepONet}; and PCNO formulates the operator
directly on point clouds~\citep{Zeng2025PCNO}. In automotive
aerodynamics specifically, GA-Field predicts vehicle fields from
geometry-aware features~\citep{Zheng2026GAField} and MARIO pursues
scalable prediction with neural fields~\citep{Catalani2025MARIO}.
These approaches incorporate geometry through latent
representations, spatial neighborhoods, or attention. In the two
backbones we build on, however, neither the retrieval step nor the
token-wise transformation is conditioned on wall proximity: AB-UPT
scores query--anchor pairs by learned feature similarity without
reference to their Euclidean separation, Transolver assigns points to
physical-state slices from token features alone, and both then apply
one shared feed-forward map to every token. We retain these backbones
unchanged and modify only those two sites.

\paragraph{Position- and distance-aware attention.}
Biasing attention by how far apart two tokens are is well established
in sequence models. T5 adds a head-specific learned scalar for each
bucketed relative-position offset~\citep{Raffel2020T5}, rotary
embeddings rotate queries and
keys so that their inner product depends on relative
position~\citep{Su2024RoPE}, and ALiBi adds a penalty that grows
linearly with distance~\citep{Press2022ALiBi}. All three are defined
over token indices in a sequence and act on the pairwise score.
DA-CA differs on both counts. Its conditioning variable is a physical
coordinate of a single token, the wall distance, rather than a
relative position between two of them, and it acts on the query
representation rather than on the score, so that wall proximity
changes what a token retrieves rather than only how strongly it
couples to a given anchor.

\paragraph{Conditional modulation and geometry-aware computation.}
Making a network's computation depend on auxiliary information is a
standard tool. FiLM modulates features by an affine map from a
conditioning signal~\citep{Perez2018FiLM}, and DiT regresses adaptive
LayerNorm scale and shift from timestep and class
embeddings~\citep{Peebles2023DiT}.
In CFD surrogates, geometry-derived signals most often enter as input
features, with signed distance and positional encodings folded into a
learned geometry representation~\citep{Ranade2025DoMINO}. Others let
geometry decide which points may interact at all, as in Erwin's
ball-tree neighborhoods~\citep{Zhdanov2025Erwin}. These approaches
condition feature representations or interaction topology. DA-CA
adopts the same affine mechanism as FiLM and DiT, but differs in what
drives it and in what it acts on. The conditioning variable is neither
a learned embedding nor a training-time index such as a diffusion
timestep, but a prescribed physical coordinate of the token itself,
the wall distance; and the modulation is applied to the query of a
cross-attention layer rather than to the residual stream, so that it
changes what a token retrieves from another branch rather than how its
own features are normalized. SVMoE uses the same class of prescribed
geometry-derived inputs, but as router inputs that select among
branch-specific token transformations rather than as modulation
parameters.

\paragraph{Mixture of experts and physical routing.}
Mixture-of-experts architectures distribute computation across
several experts to add capacity without evaluating every expert
independently for every
token~\citep{Fedus2022Switch,Puigcerver2024SoftMoE}. Sparse models
such as Switch Transformer send each token to a small number of
experts~\citep{Fedus2022Switch}, whereas Soft MoE replaces discrete
assignment with differentiable weighted token-expert
interactions~\citep{Puigcerver2024SoftMoE}. In aerodynamic
surrogates, a recent example operates at the level of whole models:
the predictions of three pretrained surrogates are combined by a
gating network that learns a spatially varying weight from features,
with entropy regularization keeping the experts
balanced~\citep{Nabian2025PhysicsNeMoMoE}. SVMoE differs in two
respects. The mixture sits inside the feed-forward layer rather than
on top of separately trained models, and its router, though learned,
is driven only by prescribed inputs rather than by
latent token features: wall distance in
the volume, a local geometric descriptor on the surface. We adopt a
pure-gate formulation without a residual
dense term; the motivation and its effect on expert specialization
are discussed in Section~\ref{sec:svmoe}.

\paragraph{Automotive aerodynamic benchmarks.}
We evaluate on two complementary DrivAer-based benchmarks.
DrivAerML~\citep{Ashton2024DrivAerML} provides high-fidelity
hybrid RANS-LES fields for 500 parametrically morphed variants of the
DrivAer notchback geometry,
suitable for direct per-cell supervision, while
DrivAerNet++~\citep{Elrefaie2024DrivAerNetPP} spans multiple body
types and enables out-of-distribution evaluation across geometric
families.

\section{Backbones and Experimental Setup}
\label{sec:overview}

\subsection{Backbones}
\label{sec:backbone}

We apply the two mechanisms to two backbones, AB-UPT and
Transolver-3, which implement the retrieval and transformation steps
in different ways.

\paragraph{AB-UPT.}
AB-UPT~\citep{Alkin2025ABUPT} separates the encoding of the shape
from the prediction of the fields by organizing the network into a
geometry branch, a surface branch, and a volume branch, which
exchange information through cross-attention so that a point in the
volume can draw on the surface near it. Applying this attention to
every point of an industrial mesh is not feasible, since its cost
grows quadratically with the number of points and a single case here
contains on the order of $10^8$ cells. AB-UPT avoids this by
selecting a small subset of points from each branch, the anchor
tokens, and computing self-attention only among them; every remaining
point then obtains its prediction by cross-attending to those
anchors. The quadratic cost is therefore paid over the anchor set
rather than over the full mesh, while predictions can still be made
anywhere on the original mesh, and it is this design that makes
training on full-resolution automotive meshes practical on a single
GPU. The retrieval site is thus the cross-attention block, and the
transformation site the feed-forward layer that follows it; stacking
several such blocks yields the final prediction of surface pressure,
wall shear stress, volume pressure, and volume velocity. Position
enters through a rotary positional encoding~\citep{Su2024RoPE}, which
rotates the queries and keys by an angle determined by their coordinates,
so that every score already carries the relative position of the two
points it compares.

\paragraph{Transolver and Transolver-3.}
Transolver~\citep{Wu2024Transolver} reaches the same goal by a
different compression. Instead of subsampling the mesh, it learns a
soft assignment of every point to a small number of physical-state
slices, forms one token per slice as the corresponding weighted mean
of point features, applies self-attention among the slice tokens, and
then recovers each point's features as the weighted combination of
those tokens under the same assignment. The retrieval site is
therefore the point--slice assignment, which plays the role that
cross-attention to anchors plays in AB-UPT: it is the step at which a
point decides what part of the compressed representation it reads.
The transformation site is again the feed-forward layer of each
block. Transolver++~\citep{Luo2025TransolverPP} and
Transolver-3~\citep{Zhou2026Transolver3} carry this formulation to
million- and industrial-scale geometries; we adopt Transolver-3, the
most recent member of the family, rather than the original
Transolver, as our second baseline.

\paragraph{Common structure.}
Both backbones follow the same two steps: each point first gathers
information from a small set of compressed tokens, and the result
then passes through a feed-forward network shared by all points. In
neither backbone does either step take into account where the point
sits in the flow. Sections~\ref{sec:daca} and~\ref{sec:svmoe}
address the two steps in turn. We develop and analyze both mechanisms on AB-UPT, where the
attention profile can be inspected directly, and report the same
mechanisms applied to Transolver-3 in
Section~\ref{sec:svmoe-results}.

All AB-UPT experiments in this paper use the surface and volume
branches and omit the geometry branch. Omitting it does change the
results slightly, but in preliminary runs on DrivAerML the difference
in field error was small compared with the effects studied here,
while the branch added parameters and training time. We therefore exclude
it and keep every comparison on the same two-branch configuration,
which applies identically to the unconditioned baseline and to the
proposed model.

\subsection{Data scaling of the unconditioned baseline}
\label{sec:data-scaling}

\begin{figure}[!htbp]
    \centering
    \begin{subfigure}[c]{0.45\linewidth}
        \centering
        \includegraphics[width=\linewidth]{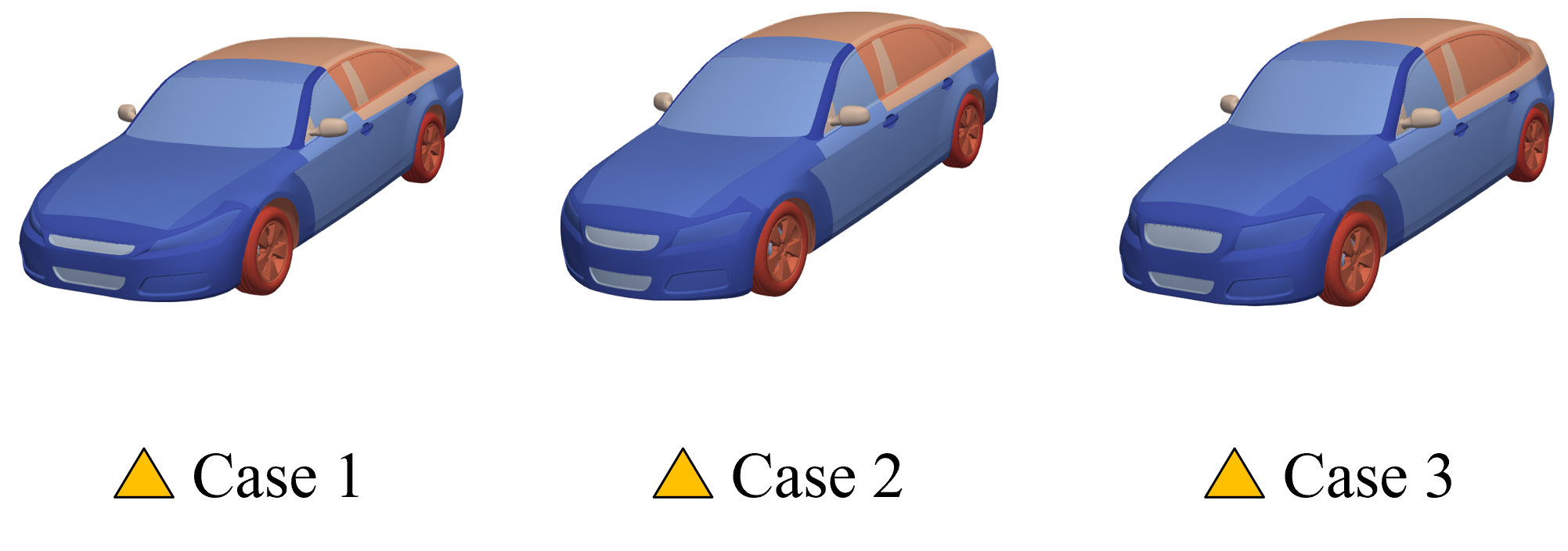}\\[0.6em]
        \small
        \begin{tabular}{@{}lc@{}}
            \toprule
            Setting & Value \\
            \midrule
            GPU                  & RTX 3090 (24 GB) \\
            Optimizer            & Lion \\
            Learning rate        & $5 \times 10^{-5}$ \\
            Batch size           & 1 \\
            Latent dimension     & 192 \\
            Train / test split   & 50 / 50 \\
            Training time        & $\sim$11 h \\
            \bottomrule
        \end{tabular}
        \caption{The three held-out DrivAerML test cases (top),
        obtained by parameterized deformation of the reference
        notchback geometry, and the training configuration shared
        across all data-scaling runs (bottom).}
        \label{fig:scaling-setup}
    \end{subfigure}
    \hfill
    \begin{subfigure}[c]{0.50\linewidth}
        \centering
        \includegraphics[width=\linewidth]{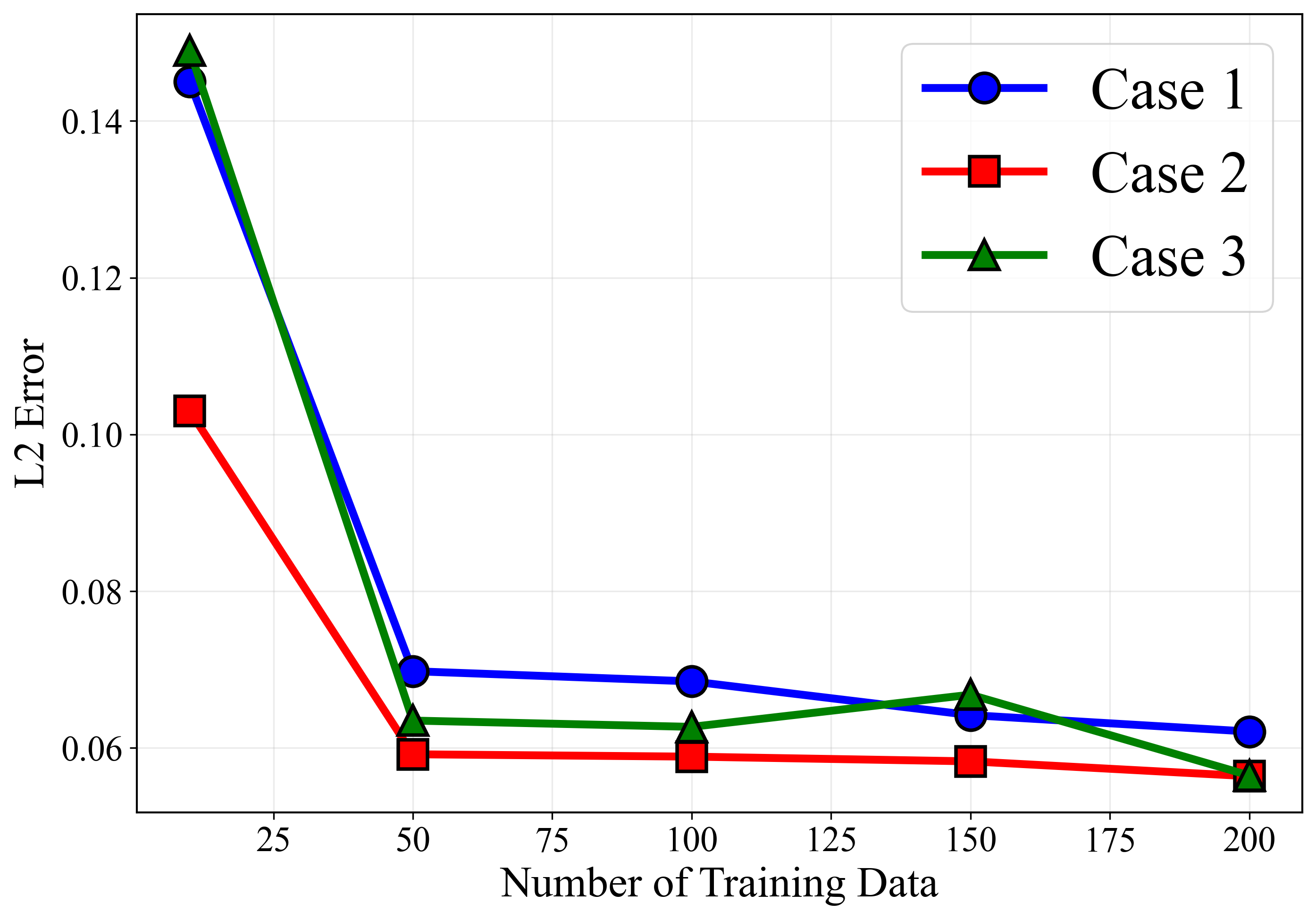}
        \caption{Relative $L_2$ error on each of the three test
        cases as a function of training-set size.}
        \label{fig:scaling-curve}
    \end{subfigure}
    \caption{Data-scaling behavior of the unconditioned baseline on
    DrivAerML. The baseline is trained at several training-set sizes
    and evaluated on three held-out test cases
    (\subref{fig:scaling-setup}, top), obtained by parameterized
    deformation of the reference notchback geometry. All runs share
    the training configuration in (\subref{fig:scaling-setup},
    bottom), with the number of epochs adjusted so that every
    training-set size consumes the same wall-clock time. Panel
    (\subref{fig:scaling-curve}) plots the relative $L_2$ error of
    each test case against the number of training samples.}
    \label{fig:scaling}
\end{figure}

A natural first response to the limitations identified above is to
ask whether more training data would suffice. We retrained the
unconditioned baseline on DrivAerML at training-set sizes
$N_{\text{train}} \in \{10, 50, 100, 150, 200\}$ under a fixed
compute budget, and evaluated each model on three held-out cases
(Figure~\ref{fig:scaling}). The relative $L_2$ error saturates near
$N_{\text{train}}=50$: quadrupling the training set from there to
200 samples leaves all three test cases largely unchanged.
Section~\ref{sec:svmoe-results} revisits the data axis directly,
retraining both backbones, with and without the proposed components,
on 300 training samples.

We read this saturation as an indication that, under the training
budget relevant to our setting, the inductive bias of the surrogate
limits accuracy more than the amount of training data does. A few
tens of samples are already sufficient to fit the dominant statistics
of the field, yet insufficient for the model to recover the regime
structure imposed by wall-bounded turbulence. The comparison holds
wall-clock time fixed, so the later configurations also receive fewer
optimization steps per sample; longer schedules or a substantially
larger high-fidelity dataset may eventually narrow the gap, and the
saturation should be read as a statement about this budget rather
than about the data axis in general.
The remainder of this paper therefore takes an
\emph{architecture-centric} stance: rather than enlarging the
training set, we modify two specific computational sites of the
transformer surrogate, the attention layer
(Sec.~\ref{sec:daca-mechanism}) and the feed-forward network
(Sec.~\ref{sec:svmoe}), so that geometry-derived physical signals
enter the computation explicitly at both stages.

\section{Distance-Aware Cross-Attention (DA-CA)}
\label{sec:daca}

\begin{figure}[!htbp]
    \centering
    \includegraphics[width=\linewidth]{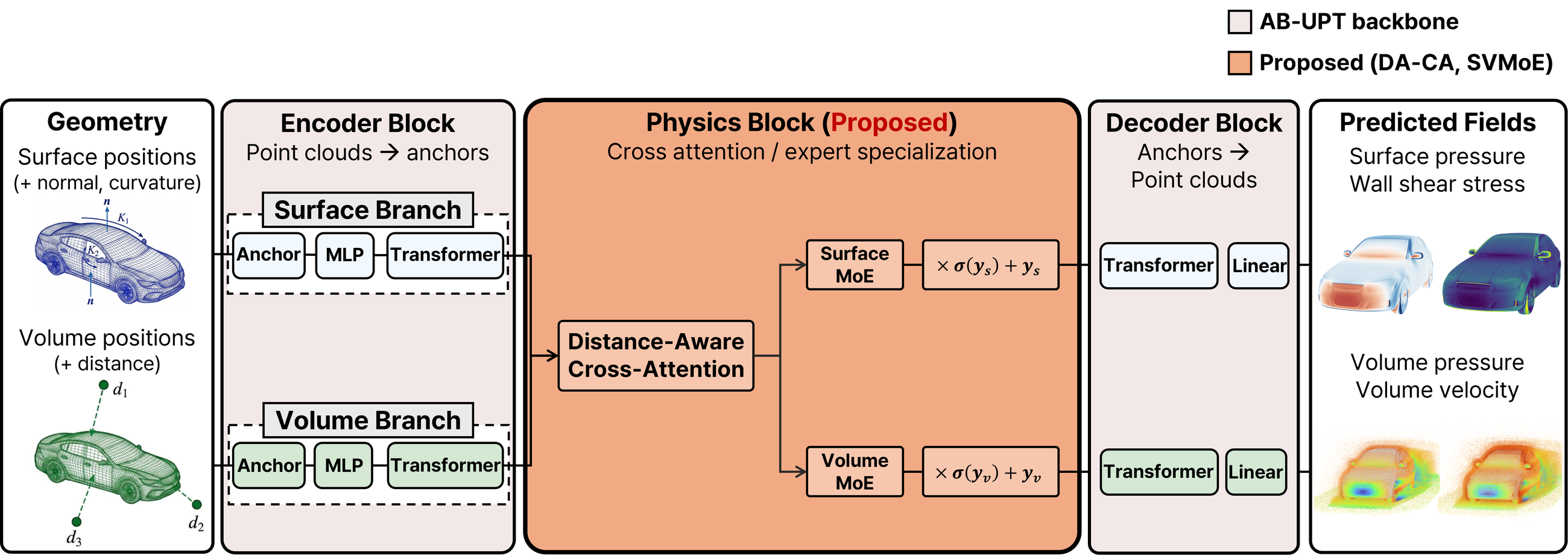}
    \caption{Architecture of the proposed model. The encoder
    compresses the surface and volume meshes into two sets of anchor
    tokens, the physics block exchanges information between the two
    branches and transforms each token, and the decoder maps the
    result back to the mesh points. The two proposed components,
    shaded in orange, both sit inside the physics block: DA-CA
    replaces the cross-attention layer and SVMoE the shared
    feed-forward layer. The rest is the AB-UPT
    backbone~\citep{Alkin2025ABUPT}, unchanged. The figure shows the
    AB-UPT instantiation; on Transolver-3 the same two components
    occupy the point--slice assignment and the block feed-forward
    layer respectively.}
    \label{fig:architecture-overview}
\end{figure}

The proposed model adds one component at each of the two sites
identified in Section~\ref{sec:backbone}: DA-CA at the
cross-attention layer, treated in this section, and SVMoE at the
feed-forward layer (Section~\ref{sec:svmoe}).
Figure~\ref{fig:architecture-overview} gives the complete picture.
Both are local changes that leave the surrounding anchor structure,
the embedding dimensions, and the optimization protocol unchanged,
so the contribution of each can be assessed in isolation, and we present
them in this order because the attention-side diagnosis motivates the
feed-forward intervention that follows. The description is given in
terms of the AB-UPT cross-attention block, on which the mechanisms
were developed; it applies to Transolver-3 with the point--slice
assignment in place of the query--anchor score, since in both cases
the modulation acts on the query-side projection of the point before
any comparison is made.

This section identifies the attention-side limitation of the
unconditioned baseline and introduces DA-CA as a targeted remedy,
together with empirical validation on DrivAerML. We will find
that DA-CA improves the near-wall and side regions but exposes a
structural limit in the wake (Sec.~\ref{sec:daca-results}),
motivating the feed-forward intervention of
Section~\ref{sec:svmoe}.

\subsection{Limitations of vanilla attention}
\label{sec:daca-motivation}

\paragraph{Cross-attention governs accuracy.}
Our intervention modifies the cross-attention layer, so we first
establish that cross-attention is in fact the component that governs
accuracy in this backbone. We compare three attention configurations
that hold the total number of attention blocks fixed at twelve but
vary how those blocks are split between self-attention and
cross-attention: the baseline split of six self-attention and six
cross-attention blocks, a self-only variant in which all twelve are
self-attention, and a cross-only variant in which all twelve are
cross-attention. Table~\ref{tab:attn-ablation} reports the resulting
errors. Removing cross-attention
entirely (self-only) degrades every
field quantity substantially: the volume velocity error rises from
7.74\% to 9.73\%, whereas removing self-attention (cross-only)
leaves accuracy close to the baseline and even slightly improves the
surface pressure prediction. The inter-branch information carried by
cross-attention, through which a volume query gathers information
from the surface anchors, is therefore the dominant contributor to
accuracy, while self-attention plays a secondary role. This is what
makes the cross-attention layer the right place to add a physical
prior: if cross-attention governs accuracy, conditioning it on wall
proximity is where such a prior should have the largest effect.
Why the unconditioned kernel nonetheless fails to exploit the regime
structure of wall-bounded flow follows from three limitations.

\begin{table}[H]
\centering
\caption{Attention-configuration ablation on DrivAerML (relative
$L_2$). All three configurations use twelve attention blocks in
total, redistributed between self- and cross-attention: six and six
in the baseline split, twelve of a single type in each variant. The
lowest error in each row is shown in bold.}
\label{tab:attn-ablation}
\small
\begin{tabularx}{\linewidth}{@{}lYYY@{}}
\toprule
& \shortstack{Baseline\\(self + cross)} & Self-only & Cross-only \\
\midrule
Surface $\Cp$    & 6.61\%  & 8.00\%  & \textbf{6.16\%} \\
Surface WSS      & \textbf{10.56\%} & 12.88\% & 10.86\% \\
Volume $\Cp$     & \textbf{5.65\%}  & 7.21\%  & 5.90\% \\
Volume Velocity  & \textbf{7.74\%}  & 9.73\%  & 8.05\% \\
\bottomrule
\end{tabularx}
\end{table}

\paragraph{Limitation 1: the attention kernel is blind to wall
proximity.}
The cross-attention used in the baseline computes, for a query
token with feature vector $h_i$ and a key token with feature
vector $h_j$,
\begin{equation}
    \alpha_{ij} \;=\; \operatorname{softmax}_j\!\left(
        \frac{(W_Q h_i)^\top (W_K h_j)}{\sqrt{d_k}}
    \right),
    \label{eq:vanilla-attn}
\end{equation}
where $h_i$ and $h_j$ already encode positional, geometric, and
physical features sampled from the surface and volume meshes.
The wall distance $\dw$, however, is not an explicit input to the
kernel: the bilinear form $(W_Q h_i)^\top (W_K h_j)$ sees only
learned features, and any wall-dependent behavior must therefore be
\emph{reconstructed} by the learned projections $W_Q, W_K$ from
whatever coordinate information the positional encoding has placed
inside $h$.

In principle, this reconstruction is available: the positional
encoding carries the 3D coordinates of each token in a sinusoidal
form that the model could learn to compare. In
practice, as we show empirically in
Section~\ref{sec:daca-results}
(Figure~\ref{fig:daca-attn-dist}), the resulting cross-attention
weights remain almost flat as a function of pairwise distance,
even for queries whose nearest anchors lie within the same
physical regime. Whatever distance information $h$ carries, the
baseline projections do not extract it in a way that organizes
attention along the wall-distance axis of the flow. We do not
claim that this reconstruction is impossible in principle, only
that under the data and compute budgets typical of high-fidelity
CFD surrogate training, it does not happen reliably. This motivates
supplying the wall distance to the kernel explicitly.

\paragraph{Limitation 2: unexploited stratification by wall
distance.}
The fields we are asked to predict are constrained by the structure
of wall-bounded turbulence. An attached boundary layer successively
traverses a viscous sublayer ($y^+ \lesssim 5$), a buffer region, a
logarithmic layer, and an outer layer before recovering toward the
free stream \citep{Pope2000Turbulent,Schlichting2017BoundaryLayer},
so the wall-normal coordinate is a primary organizing axis, even
though the broader field also exhibits separation, recirculation, and
geometry-induced complexity that wall distance alone does not
capture. A surrogate that does not condition on $\dw$ ignores that
axis and is left to recover the structure from raw coordinates
through optimization alone.

\begin{figure}[!htbp]
    \centering
    \begin{subfigure}[b]{0.55\linewidth}
        \centering
        \includegraphics[width=\linewidth]{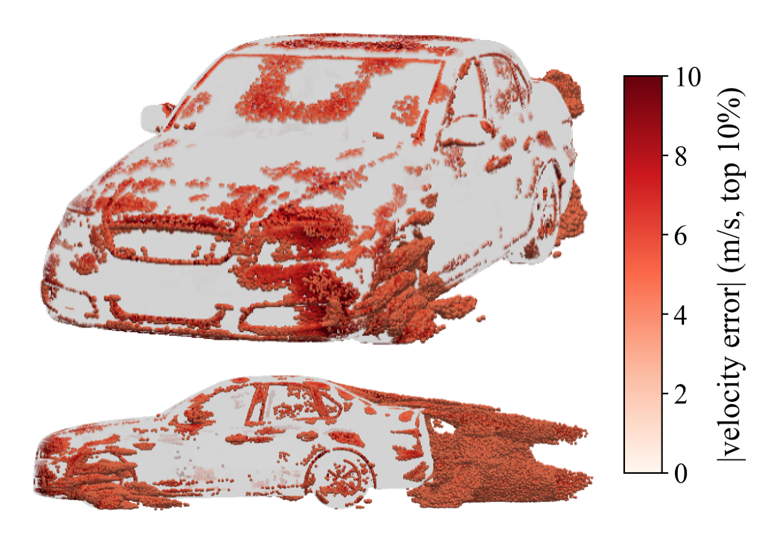}
        \caption{Spatial location of the top 10\% highest-error
        points (front and side views).}
        \label{fig:error-spatial}
    \end{subfigure}
    \hfill
    \begin{subfigure}[b]{0.42\linewidth}
        \centering
        \includegraphics[width=\linewidth]{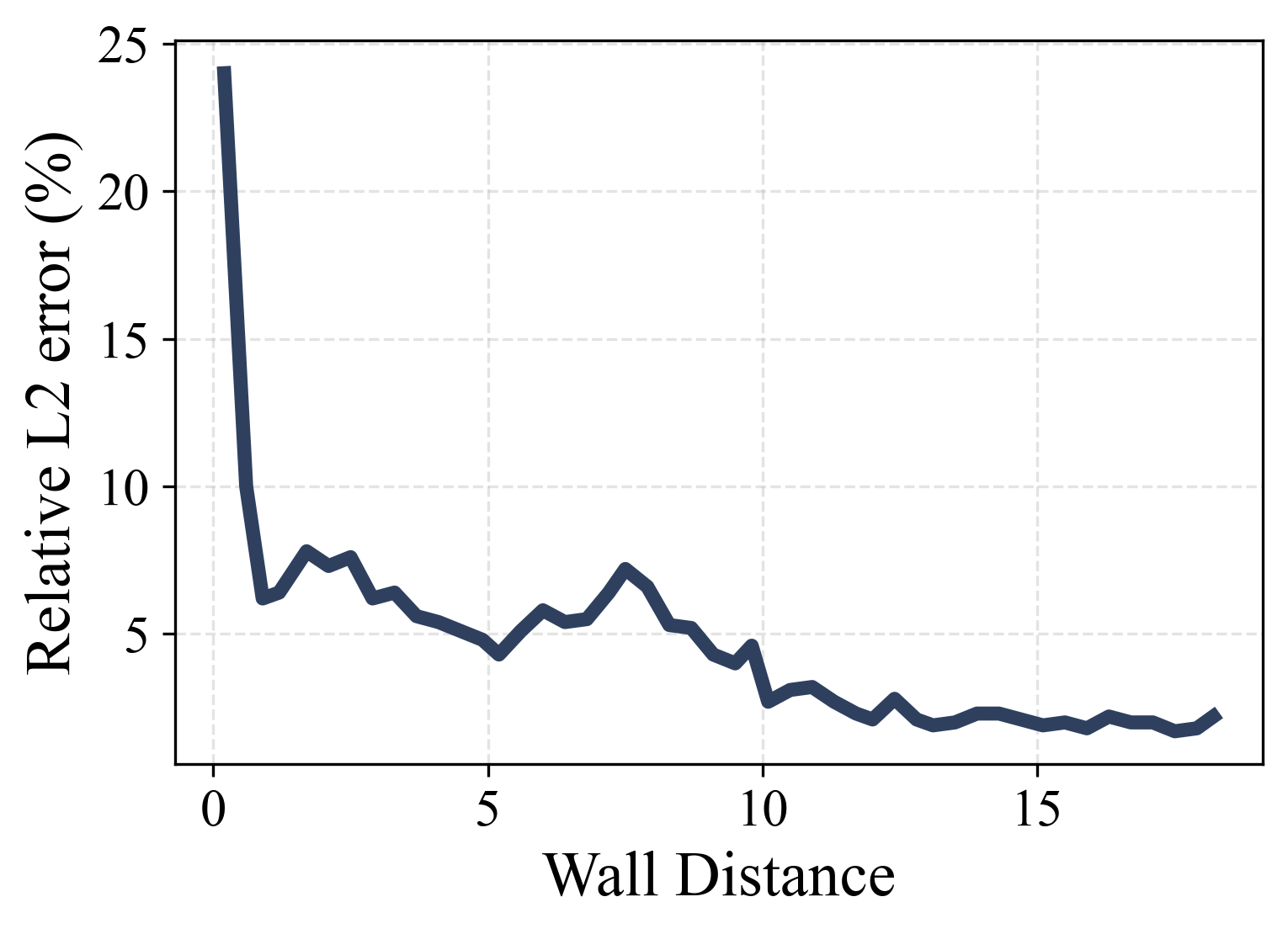}
        \caption{Relative $L_2$ error of the volume velocity field
        as a function of $\dw$.}
        \label{fig:error-vs-dw}
    \end{subfigure}
    \caption{Error localization in the unconditioned baseline on a
    DrivAerML validation sample. The highest-error points
    (\subref{fig:error-spatial}) cluster in the near-wall band and
    rear wake, and the error decays monotonically with wall
    distance (\subref{fig:error-vs-dw}). Both views place the error
    in the same part of the domain, the thin band at small $\dw$.}
    \label{fig:error-loc}
\end{figure}

\paragraph{Limitation 3: localization of prediction error.}
The first two limitations are structural. The third is empirical:
the error of the baseline is not spread evenly over the domain but
concentrated exactly where $\dw$ is small. Figure~\ref{fig:error-loc}
shows two complementary views of the prediction error of the
unconditioned baseline on a DrivAerML validation sample.
Panel~\subref{fig:error-spatial} plots the spatial location of the
top 10\% highest-error points, viewed from the front and the side:
these points cluster tightly in the near-wall band along the body
and in the rear wake, while the remainder of the domain, including
the bulk of the free-stream region, contributes negligibly to the
extreme tail of the error distribution.
Panel~\subref{fig:error-vs-dw} plots the average relative $L_2$
error of the volume velocity field as a function of $\dw$ for the
same model: the error is largest in the immediate vicinity of the
wall and decreases monotonically as $\dw$ grows, reaching its
smallest values in the free stream.

In the remainder of Section~\ref{sec:daca}, we use this
unconditioned baseline, trained under an identical protocol, as a
controlled point of comparison; all DA-CA results in
Section~\ref{sec:daca-mechanism} are reported against it.

\subsection{The DA-CA mechanism}
\label{sec:daca-mechanism}

DA-CA is a local modification of the cross-attention layer that makes
each volume query aware of its own wall distance $\dw$ before it
gathers information from the surface anchors, addressing all three
limitations at once: it supplies the missing quantity, measured along
the axis the flow is organized by, and varies most sharply in the
band where the error is concentrated.

\paragraph{Two ways to condition attention on wall proximity.}
There are two ways to bring this coordinate into cross-attention, and
we implemented both. Throughout, let $h_v \in \mathbb{R}^d$ and $x_v \in \mathbb{R}^3$
be the latent feature vector and the position of a volume query token,
$\dw(x_v) \in \mathbb{R}_{\geq 0}$ its wall distance, and
$\{(h_j^a, a_j)\}_{j=1}^{M}$ the feature vectors and positions of the
surface anchor tokens it attends to.

The first route converts the coordinate into a pairwise quantity, the
query--anchor separation $d_{vj} = \lVert x_v - a_j \rVert_2$, and
uses a learned per-head factor of it to rescale the attention score,
\begin{equation}
    \alpha_{vj}^{(h)} \;=\;
    \operatorname{softmax}_j\!\left(
        \frac{\big(W_Q^{(h)} h_v\big)^\top \big(W_K^{(h)} h_j^a\big)}
             {\sqrt{d_k}}
        \;\cdot\; s_h(d_{vj})
    \right),
    \qquad
    s_h(d) \;=\; b_h + \exp\!\left(-\frac{d}{\sigma_h}\right),
    \label{eq:daca-pairwise}
\end{equation}
where $\sigma_h$ sets the interaction range of head $h$ and the floor
$b_h > 0$ keeps distant anchors active. The factor rescales the
magnitude of each logit as a function of separation, so the weight a
query places on an anchor is modulated by how far apart the two
lie.

The second route keeps $\dw$ as the single-token property it is and
uses it to reshape the query before the score is formed, so that how
deep a token sits in the boundary layer changes what it looks for in
the surface. We encode the wall distance on a logarithmic,
multi-scale basis,
\begin{equation}
    \phi(\dw) \;=\;
    \Big[\,\log \dw,\;
    \big\{\sin(\omega_k \log \dw),\,
          \cos(\omega_k \log \dw)\big\}_{k=1}^{K}\,\Big],
    \qquad \omega_k = 2^{\,k-1}\pi,
    \label{eq:daca-fourier}
\end{equation}
map it to per-head scale and shift parameters through a small
multilayer perceptron,
\begin{equation}
    \big(\gamma^{(h)}(\dw),\,\beta^{(h)}(\dw)\big)_{h=1}^{H}
    \;=\; \operatorname{MLP}\!\big(\phi(\dw)\big),
    \label{eq:daca-adaln-params}
\end{equation}
and modulate the per-head query with them,
\begin{equation}
    \tilde{Q}^{(h)}_v
    \;=\; \big(1 + \gamma^{(h)}(\dw)\big)\,\odot\, W_Q^{(h)} h_v
    \;+\; \beta^{(h)}(\dw),
    \qquad
    \alpha_{vj}^{(h)}
    \;=\; \operatorname{softmax}_j\!\left(
        \frac{\big(\tilde{Q}^{(h)}_v\big)^\top\!\big(W_K^{(h)} h_j^a\big)}
             {\sqrt{d_k}}
    \right).
    \label{eq:daca-adaln}
\end{equation}
Here $\odot$ is the element-wise product over the head's feature
channels, so $\gamma^{(h)}$ and $\beta^{(h)}$ act as a
wall-distance-dependent gain and bias on the query. The keys and
values, and hence the surface representation, are left unchanged: the
modulation acts only on the branch that reads from the surface, and
the anchor structure, the embedding dimensions, and the softmax are
all untouched.

The two differ in where the coordinate enters. In
Eq.~\eqref{eq:daca-pairwise} it multiplies the score, once per
query--anchor pair; in Eq.~\eqref{eq:daca-adaln} it acts on the query
itself, before any anchor is involved. We adopt the second for two
reasons.

The first concerns which quantity is measured. The wall distance is
the coordinate along which the boundary layer is organized, whereas
the query--anchor separation says only how far this point happens to
be from the particular anchor it is compared to. Two points at the
same depth can receive very different separations depending on which
anchors were sampled nearby, so the separation is a property of the
discretization rather than of the flow: it changes with the mesh and
with the anchor sampling while $\dw$ does not. Conditioning on $\dw$
ties the model to the physics rather than to one geometry, which is
what allows the same conditioning to carry over in
Section~\ref{sec:ood}.

The second concerns where the quantity acts. A scalar factor can only
make the query longer or shorter, leaving its direction untouched, so
it changes how much weight a point gives to near and far anchors but
not what the point is looking for. The modulation acts on each
component of the query separately and therefore turns it, so a point
buried in the boundary layer and a point in the free stream can read
the same piece of surface for different things. This matters because
a factor that decays with distance can only express the preference
that closer anchors matter more, which is true near the wall and
beside the body but says nothing about the wake.

\paragraph{Relation to the positional encoding.}
As noted in Section~\ref{sec:backbone}, the backbone already applies
a rotary positional encoding to the queries and keys, and both it and
Eq.~\eqref{eq:daca-fourier} use sinusoids, so it is worth saying why
the two do not overlap. The rotary encoding rotates $Q$ and $K$ by a
fixed, non-learned angle determined by the raw coordinates; it preserves
their norms and makes the score depend on the relative position of the
two tokens. The modulation here is learned, acts on the query alone,
and rescales and shifts its components rather than rotating them. More
importantly, the two carry different information: $\dw$ is not a
coordinate but a derived quantity that requires knowing where the
nearest wall lies, and is therefore not recoverable from the
coordinates of a single token.
Table~\ref{tab:daca-signals} places the three signals side by side.

\begin{table}[!htbp]
\centering
\caption{What each mechanism supplies to the cross-attention layer.
The positional encoding belongs to the backbone and is always active;
the other two are alternative ways of adding wall proximity on top of
it, of which we adopt the second.}
\label{tab:daca-signals}
\small
\begin{tabularx}{\linewidth}{@{}lYY@{}}
\toprule
& Quantity supplied & Where it acts \\
\midrule
Positional encoding (backbone)  & raw coordinates, hence relative position & rotates $Q$ and $K$ \\
Pairwise scaling (considered)   & distance to a given anchor & scalar factor on the score \\
\textbf{Wall conditioning (adopted)} & distance to the wall & per-component modulation of $Q$ \\
\bottomrule
\end{tabularx}
\end{table}

\paragraph{Design choices.}
Two properties of
Eqs.~\eqref{eq:daca-fourier}--\eqref{eq:daca-adaln} deserve comment.

1) \emph{Logarithmic, multi-scale encoding.}
The length scales of an external wall-bounded flow span at least
three orders of magnitude, from the viscous sublayer to the far
field, and a linear treatment of $\dw$ would resolve only one of
them. The logarithm in Eq.~\eqref{eq:daca-fourier} places these bands on a
comparable footing, and the Fourier features $\{\omega_k\}$ let the modulation
vary sharply across the near-wall bands while remaining smooth in the
far field, so that each head can specialize to a distinct range of
wall distances.

2) \emph{Zero initialization.}
The output layer of the MLP starts at zero, so training begins from
$\tilde{Q}^{(h)}_v = W_Q^{(h)} h_v$, that is, from the unconditioned
cross-attention itself. The model departs from the baseline only as
the conditioning is learned, so any improvement is attributable to
that conditioning rather than to a different starting point.

\subsection{Results and analysis}
\label{sec:daca-results}

\begin{figure}[!htbp]
    \centering
    \includegraphics[width=\linewidth]{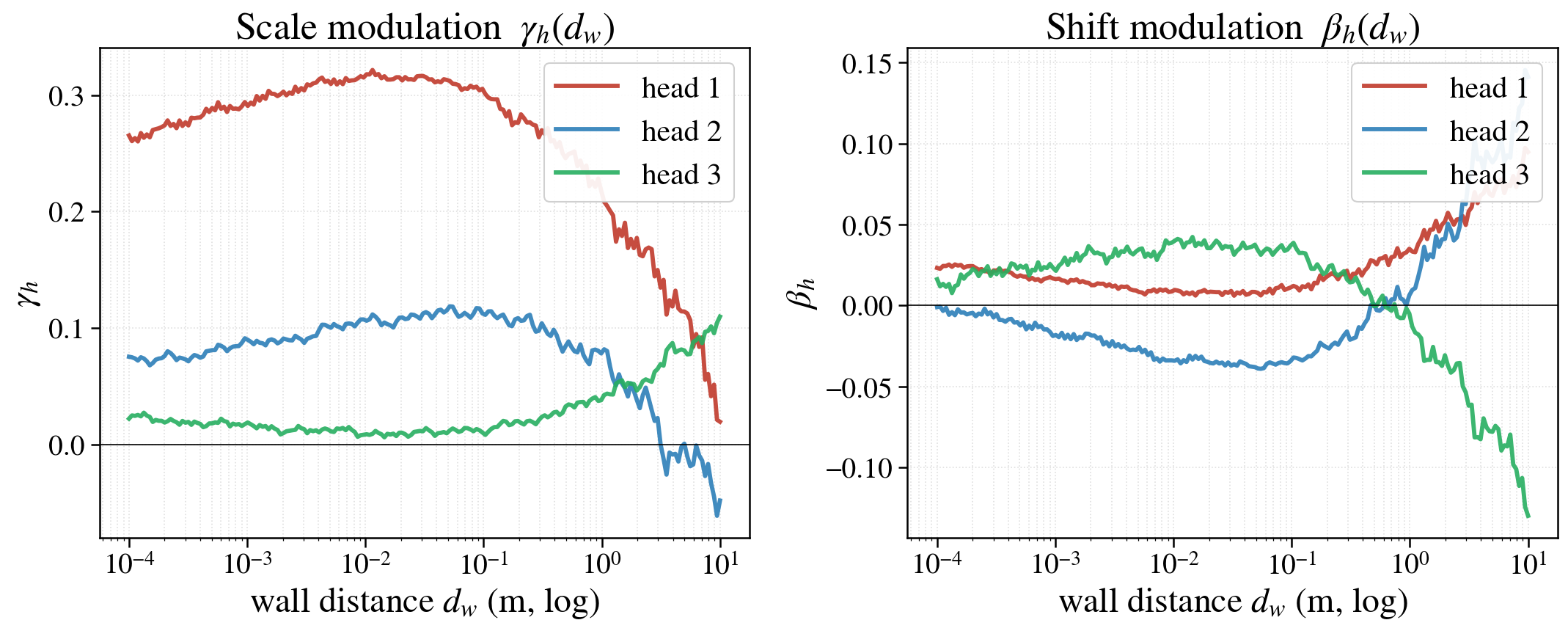}
    \caption{Learned scale $\gamma^{(h)}$ and shift $\beta^{(h)}$ of
    the DA-CA modulation against wall distance, after training on
    DrivAerML. Each cross-attention block carries three heads, and the
    curves shown are those of a single block. Both parameters are zero
    at initialization, so any structure here is learned.}
    \label{fig:daca-modulation}
\end{figure}

\paragraph{The learned modulation.}
Before looking at accuracy, we check whether the conditioner is used
at all. Because the modulation MLP is initialized to zero, DA-CA
starts as an exact copy of the unconditioned cross-attention, so any
departure from $\gamma^{(h)} = \beta^{(h)} = 0$ has to be learned.
Figure~\ref{fig:daca-modulation} plots the converged
$\gamma^{(h)}(\dw)$ and $\beta^{(h)}(\dw)$ against wall distance. The
curves move well away from zero, reaching a scale of $0.32$ at their
largest, so the conditioning is recruited rather than left inactive.
They also vary strongly with $\dw$, and they do so differently for
different heads: of the three heads in a block, two place their
largest scaling inside the near-wall band, below roughly
$10^{-1}$~m, and fall off beyond it, while the third stays small
there and grows only in the outer flow.
This is the specialization that the multi-scale encoding of
Eq.~\eqref{eq:daca-fourier} was meant to permit, and it is not
imposed: nothing in the loss asks the heads to divide the range of
wall distances between them.

\begin{figure}[!htbp]
    \centering
    \includegraphics[width=\linewidth]{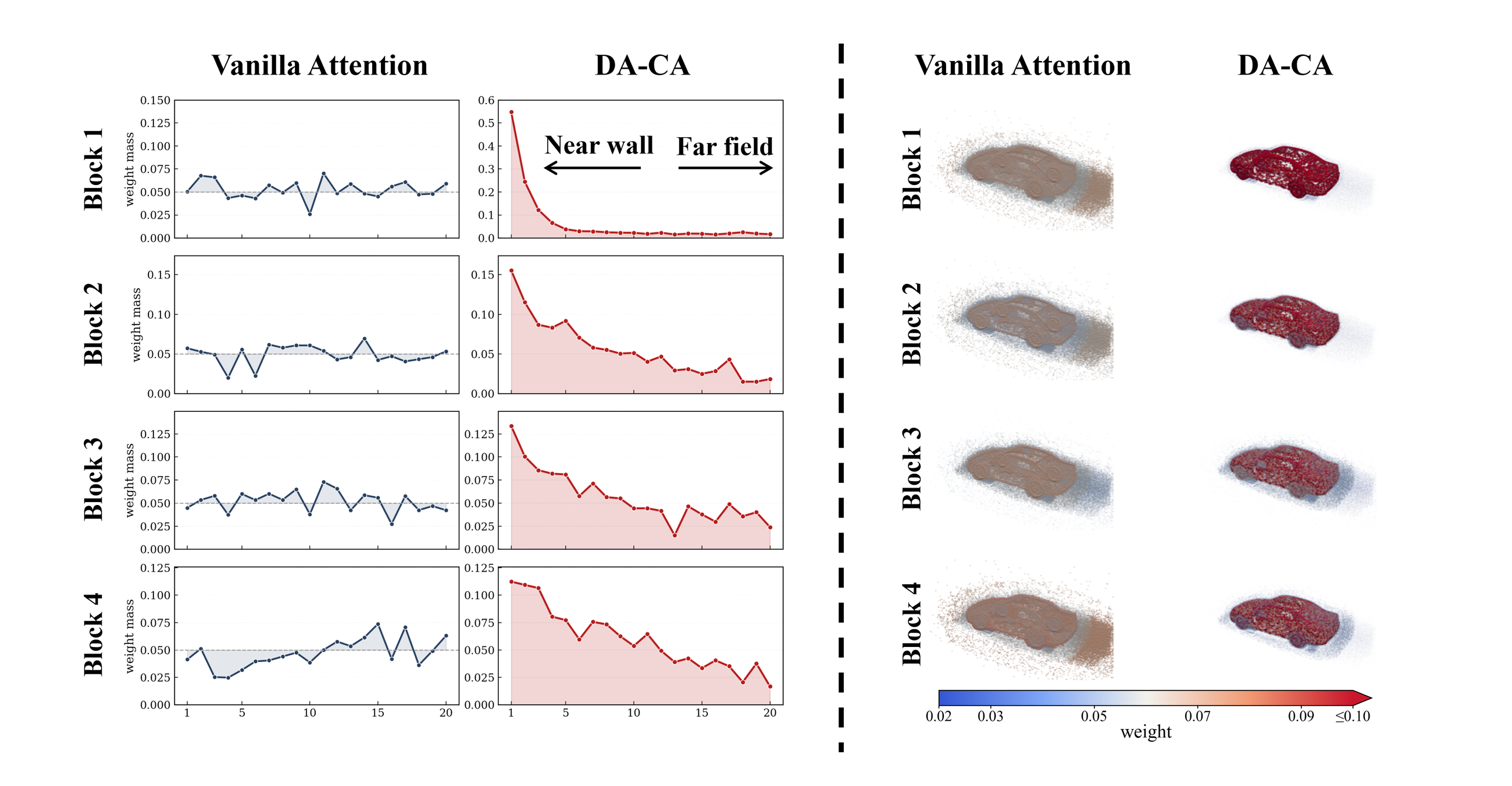}
    \caption{Cross-attention weight distribution by distance rank
    on a DrivAerML validation sample. Left: weight mass per distance
    bin, with anchors grouped into 20 bins by distance (bin 1 =
    nearest, bin 20 = farthest), for the unconditioned baseline
    (blue) and DA-CA (red), for four of the six cross-attention
    blocks. Right: the same weights visualized on the vehicle
    surface. The nearest-bin mass increases by roughly an order of
    magnitude under DA-CA.}
    \label{fig:daca-attn-dist}
\end{figure}

\paragraph{Distance-rank attention.}
DA-CA is inserted in the cross-attention blocks, leaving the
self-attention blocks unchanged, so the intervention stays localized
to the cross-attention path. To ask how the conditioning changes the
attention profile,
Figure~\ref{fig:daca-attn-dist} shows the cross-attention weight
distribution on a DrivAerML validation sample, with anchors grouped
into 20 bins by distance rank (bin 1: nearest, bin 20: farthest),
for four of the six cross-attention blocks. The unconditioned
baseline produces a nearly flat profile in every block, close to the
uniform-attention reference. DA-CA shifts mass toward nearby anchors,
producing a monotone distance-rank profile in every block shown.
DA-CA is never given the query--anchor separation, so this distance
dependence is not imposed by construction: it emerges from
conditioning each query on its own wall distance.

\begin{figure}[!htbp]
    \centering
    \includegraphics[width=\linewidth]{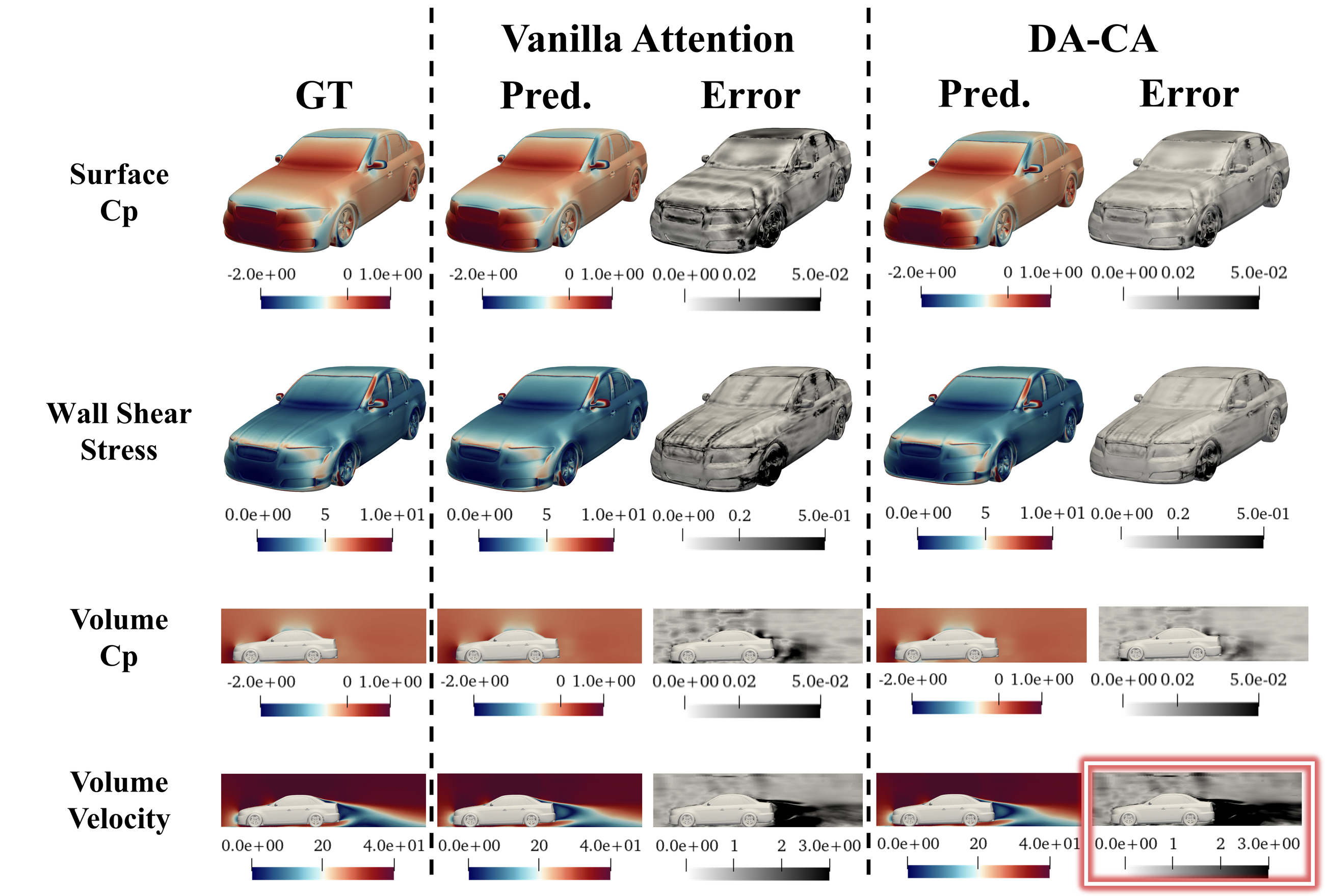}
    \caption{For each field, the baseline
    block shows ground truth, prediction, and error, while the DA-CA
    block shows its prediction and error against the same ground
    truth. DA-CA reduces surface error in the near-wall band and
    recovers the front-stagnation structure more cleanly in the
    volume.}
    \label{fig:daca-fields}
\end{figure}

\paragraph{Field-level comparison (DA-CA vs.\ baseline).}
The attention change is accompanied by lower field errors.
Figure~\ref{fig:daca-fields} compares the surface
pressure coefficient $\Cp$, wall shear stress $\WSS$, volume
pressure coefficient $\Cp$, and a
representative volume-velocity slice for the unconditioned
baseline and DA-CA on the same validation sample. DA-CA reduces
spurious surface error in the near-wall band and gives a sharper
front-bumper stagnation pattern in the representative volume slice.
Quantitatively,
Table~\ref{tab:ablation} confirms that DA-CA reduces
the volume pressure error by 10.1\% under the 50-sample training
regime.

\begin{table}[!htbp]
\centering
\caption{DrivAerML validation errors (relative $L_2$) under the
50-sample regime, averaged over the validation set. DA-CA reduces
error on all four field quantities relative to the unconditioned
baseline. Parentheses give the relative reduction in error with
respect to that baseline.}
\label{tab:ablation}
\footnotesize
\setlength{\tabcolsep}{4pt}
\begin{tabularx}{\linewidth}{@{}lYYYY@{}}
\toprule
Model & Surface $\Cp$ & WSS & Volume $\Cp$ & Velocity \\
\midrule
AB-UPT                      & 6.89\% & 11.40\% & 5.94\% & 8.23\% \\
\textbf{DA-CA}              & \textbf{6.46\%} \textbf{($-$6.2\%)} & \textbf{10.63\%} \textbf{($-$6.8\%)} & \textbf{5.34\%} \textbf{($-$10.1\%)} & \textbf{7.73\%} \textbf{($-$6.1\%)} \\
\bottomrule
\end{tabularx}
\end{table}

\begin{figure}[!htbp]
    \centering
    \begin{minipage}[c]{0.38\linewidth}
        \centering
        \begin{subfigure}[c]{\linewidth}
            \centering
            \includegraphics[width=\linewidth]{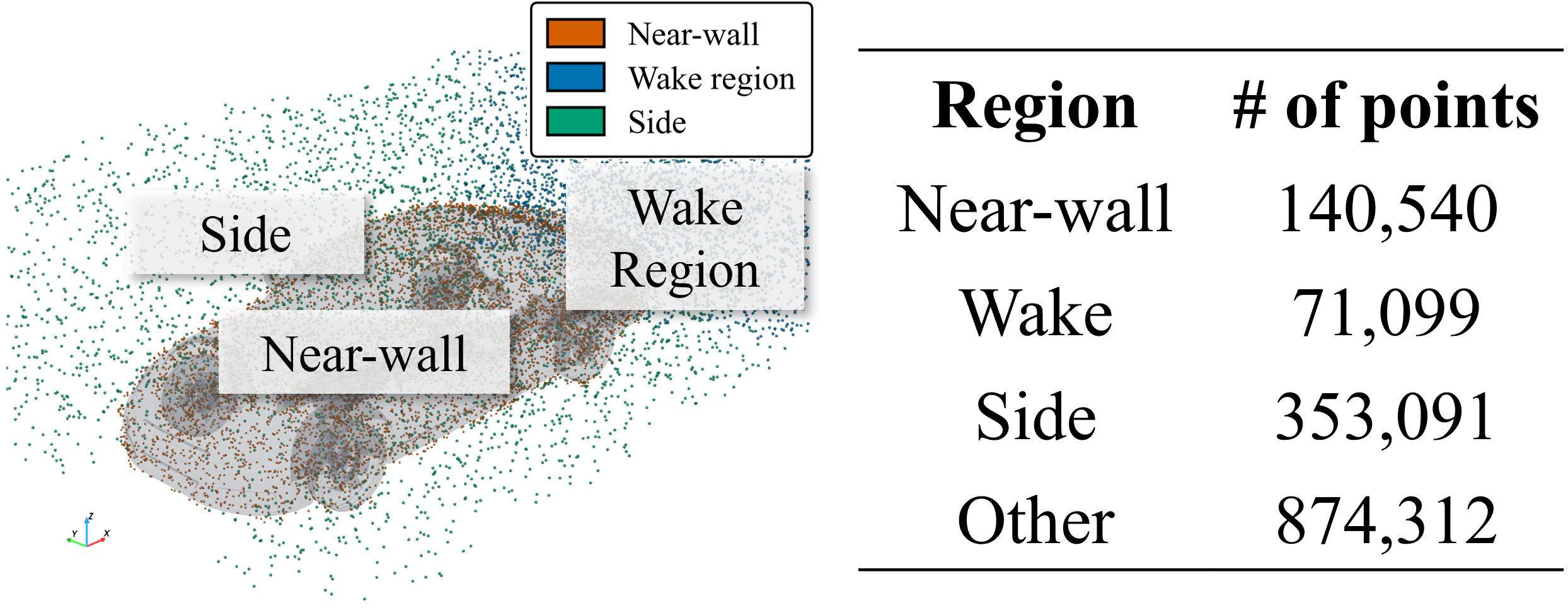}
            \caption{Spatial partitioning of the volume domain into
            near-wall, wake, and side regions, with all remaining
            points grouped as \emph{other}, and the point count of
            each.}
            \label{fig:daca-region-partition}
        \end{subfigure}\\[0.3em]
        \begin{subfigure}[c]{\linewidth}
            \centering
            \includegraphics[width=\linewidth]{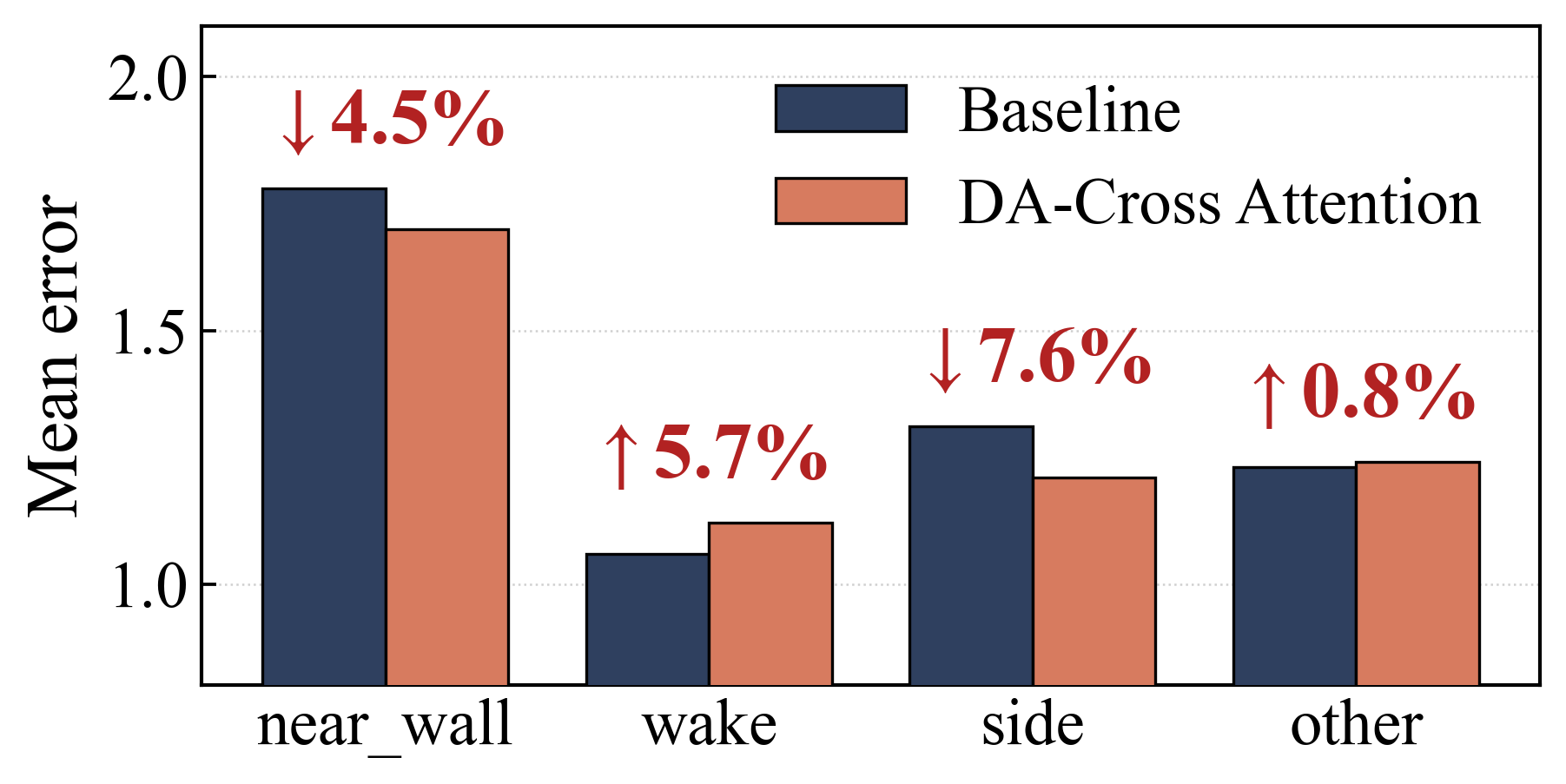}
            \caption{Mean error per region for the unconditioned
            baseline and DA-CA, with the relative change annotated
            above each pair.}
            \label{fig:daca-region-bars}
        \end{subfigure}
    \end{minipage}
    \hfill
    \begin{subfigure}[c]{0.6\linewidth}
        \centering
        \includegraphics[width=\linewidth]{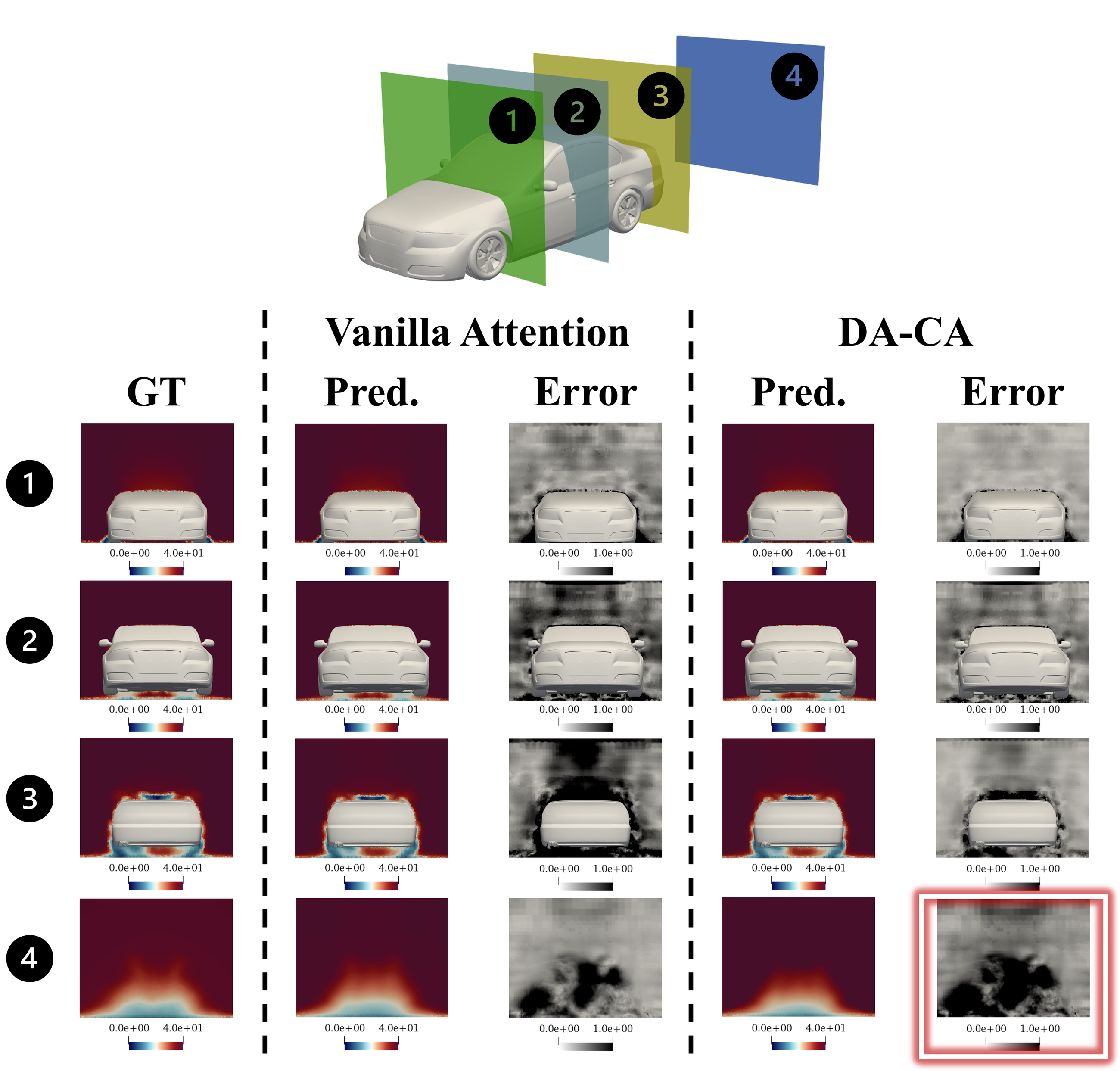}
        \caption{Velocity slices in the front-bumper region for the
        unconditioned baseline (left) and DA-CA (right).}
        \label{fig:daca-region-slice}
    \end{subfigure}
    \caption{Region-wise analysis of DA-CA on a DrivAerML
    validation sample. The panels show the spatial partition, the
    regional error bars, and representative velocity slices. DA-CA
    improves the near-wall and side regions but yields little or
    slight degradation in the wake, consistent with the cleaner
    front-bumper field.}
    \label{fig:daca-region}
\end{figure}

\paragraph{Region-wise analysis and the wake gap.}
The aggregate gain is not spatially uniform.
Figure~\ref{fig:daca-region} partitions the volume
domain into four regions based on each point's wall distance and
streamwise position: near-wall, wake, side, and all remaining points
grouped as \emph{other}. It reports the mean error in each region
together with the relative change from the baseline to DA-CA. The near-wall
and side regions improve, which is where the conditioning varies most
sharply. The wake region, however, shows little improvement
and slightly regresses on this sample. The slice visualization
suggests why: the front-bumper region is sharpened, but the wake
structure behind the vehicle remains close to the baseline
prediction.

This regional pattern follows from what DA-CA does and does not
change. Near-wall and side queries sit at small $\dw$, where the
modulation varies most sharply and where the surface beneath them
carries most of the information they need, so sharper retrieval
helps them directly. What a query retrieves is then transformed by a
feed-forward layer that DA-CA leaves untouched and that all volume
points share. As the near-wall band, which holds most of the points
and most of the error, is fitted more closely, this shared layer is
pulled toward that band, and the wake, which lies in the outer flow,
gains nothing from the sharper retrieval and loses part of the
shared transformation. The wake is therefore the exception, and it
motivates a second intervention at the feed-forward layer, which
Section~\ref{sec:svmoe} takes up.

\section{Surface-Volume Mixture-of-Experts (SVMoE)}
\label{sec:svmoe}

This section introduces SVMoE, the feed-forward component of the
proposed model. DA-CA conditions what a point gathers from the
surface; SVMoE conditions how it transforms what it gathered. The
goal is therefore to let surface tokens, and volume tokens at
different depths, use partially different nonlinear maps while
keeping the backbone unchanged.

\subsection{Motivation}
\label{sec:svmoe-motivation}

\paragraph{Motivation 1: the limit of DA-CA in the wake.}
Section~\ref{sec:daca-results} traced the residual wake error of
DA-CA to the feed-forward layer: DA-CA changes only what a token
\emph{gathers}, not how it \emph{transforms} what it gathered, and
the single transformation shared by all volume points is dominated
by the near-wall band. The residual error therefore points to the
second computational site, the feed-forward layer.

\paragraph{Motivation 2: a single FFN for two branches and three
flow regimes.}
The baseline applies a single feed-forward network $y =
\mathrm{FFN}_{\mathrm{shared}}(x)$ to every token in every block.
This shared mapping makes two simplifications. The first is that
surface and volume tokens admit the same transformation, although
surface tokens live on the two-dimensional body manifold and predict
scalar wall quantities ($\Cp$, $\WSS$) while volume tokens live in
the three-dimensional exterior domain and predict the vector
velocity and the volume pressure; the baseline maps these physically
distinct populations through identical weights. The second is that, within the volume, points at every depth admit
the same nonlinear transformation, although the mechanisms that
govern them differ: wall-normal shear dominates close to the wall,
while further out the flow is nearly inviscid except where it is
separated. Wall distance orders these cases without resolving all of
them; in particular it does not tell a wake point from a free-stream
point at the same depth, and the mixture we introduce below therefore
gives the outer band a transformation of its own rather than a
wake-specific one.

\subsection{Architecture}
\label{sec:svmoe-arch}

\begin{figure}[!htbp]
    \centering
    \includegraphics[width=\linewidth]{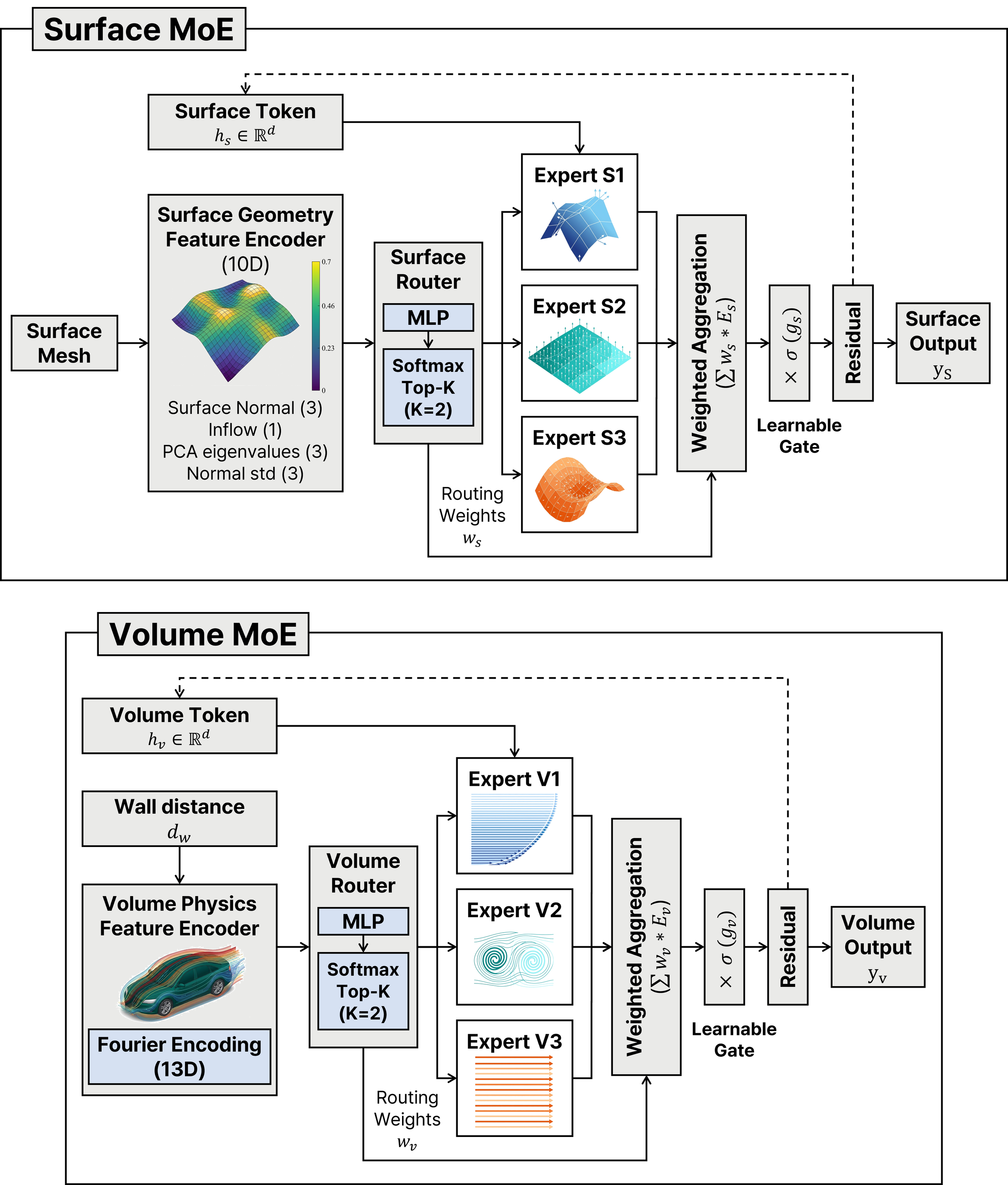}
    \caption{SVMoE architecture. Surface tokens (top) are routed by a
    10-dimensional geometric descriptor through three surface
    experts; volume tokens (bottom) are routed by Fourier-encoded log
    wall distance through three volume experts. Each branch performs
    a top-2 weighted aggregation over its experts, is scaled by a
    learned output gate, and is added back to the token through a
    residual connection. The two branches are independently
    parameterized.}
    \label{fig:svmoe-arch}
\end{figure}

\paragraph{Branch separation.}
The shared mapping is replaced by two branch-specific
transformations, $y_s = \mathrm{FFN}_s(h_s)$ for surface tokens and
$y_v = \mathrm{FFN}_v(h_v)$ for volume tokens, where $h_s$ and
$h_v$ denote the corresponding token features
(Figure~\ref{fig:svmoe-arch}). The two FFNs are
independently parameterized and trained, which ensures that
surface and volume gradients no longer overwrite each other in
the same weights; Algorithm~\ref{alg:svmoe} specifies the complete
layer. The split is by token type, not by any structure of the
surrounding network, so it applies unchanged to a backbone that
carries surface and volume points in a single stream: each token is
sent to the mixture matching its type, and the two mixtures remain
independently parameterized.

\paragraph{Volume mixture of experts.}
The volume FFN is replaced by a three-expert mixture
$E_{v,1}, E_{v,2}, E_{v,3}$ whose router depends only on the wall
distance $\dw$ (Algorithm~\ref{alg:svmoe}): a Fourier encoding
lifts the scalar $\log \dw$ to 13 dimensions, an MLP maps the
encoding to routing logits $z_v$, top-2 routing weights $w_v$
combine the experts, and the result is scaled by a scalar output
gate $g_v$ and added back to the token through a residual
connection. The gate is initialized to a small negative value so
that the block starts close to the identity and the mixture is
introduced gradually during training. The Fourier encoding lets the
router express transitions on a scale that matches the canonical
$y^+$ stratification of wall-bounded turbulence, and the top-2
selection retains a smooth interpolation across regime boundaries
instead of a hard partition.

\begin{algorithm}[!t]
\caption{SVMoE feed-forward layer, replacing the shared FFN in every
cross-attention block. The surface and volume branches are
independently parameterized, and both routers see only prescribed
physical inputs.}
\label{alg:svmoe}
\KwData{surface tokens $h_s$ with geometric descriptors
$\phi_s = \bigl[\mathbf{n},\;
\mathbf{n}\cdot\hat{\mathbf{u}}_\infty,\;
\boldsymbol{\lambda},\;
\boldsymbol{\sigma}_{\mathbf{n}}\bigr] \in \mathbb{R}^{10}$;
volume tokens $h_v$ with wall distances $\dw$}
\KwResult{transformed tokens $y_s$, $y_v$; auxiliary loss
$\mathcal{L}_{\text{aux}}$}
\tcc{volume branch: three experts $\{E_{v,k}\}_{k=1}^{3}$, routed
by wall distance}
\ForEach{volume token $(h_v, \dw)$}{
    $z_v \leftarrow \mathrm{MLP}\!\left(\mathrm{FourierEnc}(\log \dw)\right)$\;
    $w_v \leftarrow \mathrm{TopK}_2\!\left(\mathrm{softmax}(z_v)\right)$\;
    $y_v \leftarrow h_v + \sigma(g_v)\,\textstyle\sum_{k=1}^{3} w_{v,k}\, E_{v,k}(h_v)$\;
}
\tcc{surface branch: three experts $\{E_{s,k}\}_{k=1}^{3}$, routed
by local geometry}
\ForEach{surface token $(h_s, \phi_s)$}{
    $z_s \leftarrow \mathrm{MLP}(\phi_s)$\;
    $w_s \leftarrow \mathrm{TopK}_2\!\left(\mathrm{softmax}(z_s)\right)$\;
    $y_s \leftarrow h_s + \sigma(g_s)\,\textstyle\sum_{k=1}^{3} w_{s,k}\, E_{s,k}(h_s)$\;
}
\tcc{training only: load balancing on each branch}
\If{training}{
    $\mathcal{L}_{\text{aux}} \leftarrow \lambda \cdot N \cdot
    \textstyle\sum_{i=1}^{N} f_i \cdot P_i$\;
}
\end{algorithm}

\paragraph{Surface mixture of experts.}
The surface FFN is replaced by a mirror-image three-expert mixture
$E_{s,1}, E_{s,2}, E_{s,3}$ with routing weights $w_s$ and output
gate $g_s$, whose
router takes the ten-dimensional geometric descriptor
$\phi_s = \bigl[\,\mathbf{n},\;
\mathbf{n}\cdot\hat{\mathbf{u}}_\infty,\;
\boldsymbol{\lambda},\;
\boldsymbol{\sigma}_{\mathbf{n}}\,\bigr]$ at each surface token
(Algorithm~\ref{alg:svmoe}), where $\mathbf{n}\in\mathbb{R}^3$ is
the surface normal,
$\mathbf{n}\cdot\hat{\mathbf{u}}_\infty\in\mathbb{R}$ is its
alignment with the free-stream direction,
$\boldsymbol{\lambda}\in\mathbb{R}^3$ is the normalized eigenvalue
triplet $(\lambda_1,\lambda_2,\lambda_3)/\sum_i \lambda_i$ of the
covariance matrix of each point's $k$ nearest neighbors ($k=16$),
and $\boldsymbol{\sigma}_{\mathbf{n}}\in\mathbb{R}^3$ is the
component-wise standard deviation of the surface normals over those
same neighbors. The eigenvalue triplet is a PCA-based descriptor of
local surface shape rather than a differential
curvature~\citep{Weinmann2015PointCloud}: it
separates planar neighborhoods, edge-like ridges, and scattered
regions. The descriptor collects the
geometric quantities that local boundary-layer behavior depends on,
the wall orientation and its alignment with the oncoming flow, the
shape of the surrounding neighborhood, and how much the normal varies
across it, so that the router can
distinguish surface tokens in functionally distinct geometric
regimes.

\paragraph{Pure-gate mixtures.}
Both mixtures use a \emph{pure-gate} formulation,
$y = h + \sigma(g)\sum_k w_k E_k(h)$, in which the standard
identity skip connection is retained but no dense base term
$\mathrm{FFN}_{\mathrm{shared}}(h)$ is added in parallel with the
experts: such a base term provides a shortcut path through which the
optimizer can bypass the experts, suppressing the gradient signal
that drives expert specialization, whereas the pure-gate design
forces each expert to carry its share of the prediction.

\paragraph{Routing variables.}
Both routers are learned, but they see only prescribed,
geometry-derived inputs; in a small-data regime a router with access
to latent token features can memorize idiosyncrasies of the training
distribution, whereas restricting the input to $\dw$ and $\phi_s$
ties the partition to physical regimes rather than to the training
set (tested under geometric out-of-distribution evaluation in
Section~\ref{sec:ood}). The two branches take different variables
because volume tokens span several orders of magnitude in $\dw$ and
vary primarily in the wall-normal direction, whereas on the body
two-manifold $\dw$ is trivially zero and the relevant variation is
geometric, so $\phi_s$ takes its place. Both mixtures are inserted at
every cross-attention block with block-specific parameters; the
DA-CA modulation MLP of Section~\ref{sec:daca} is shared across
blocks.

\paragraph{Load-balancing loss.}
A common failure mode of top-$k$ routing is expert collapse, in
which a small subset of experts absorbs the majority of routing
mass and the remaining experts receive little or no gradient
signal. Following the Switch Transformer
formulation~\citep{Fedus2022Switch}, we augment the task loss with
the auxiliary term $\mathcal{L}_{\text{aux}}$ of
Algorithm~\ref{alg:svmoe}, applied to each branch, where $f_i$ is
the fraction of tokens for which expert $i$ is selected in the
top-$k$, $P_i$ is the mean routing probability assigned to expert
$i$, $N=3$ is the number of experts, and $\lambda=0.01$. The
auxiliary loss is minimized when routing mass is distributed
uniformly across experts, providing a soft prior against collapse
without preventing the emergence of the physical partition we
observe in Section~\ref{sec:svmoe-results}.

\subsection{Results}
\label{sec:svmoe-results}

We evaluate the full proposed model (DA-CA + SVMoE) on the same
DrivAerML protocol used in Section~\ref{sec:daca}, with 50
training samples and a matched compute budget. Unless stated
otherwise, all results in this section use both
branch-specific mixtures (surface and volume) with DA-CA enabled,
and are reported on the AB-UPT instantiation; the Transolver-3
instantiation is taken up in the 300-sample experiment below.
We report the unsupervised emergence of physically interpretable
expert partitions and the resulting improvement in predicted field
accuracy over the unconditioned baseline.

\begin{figure}[!htbp]
    \centering
    \begin{subfigure}[t]{0.49\linewidth}
        \centering
        \includegraphics[width=\linewidth]{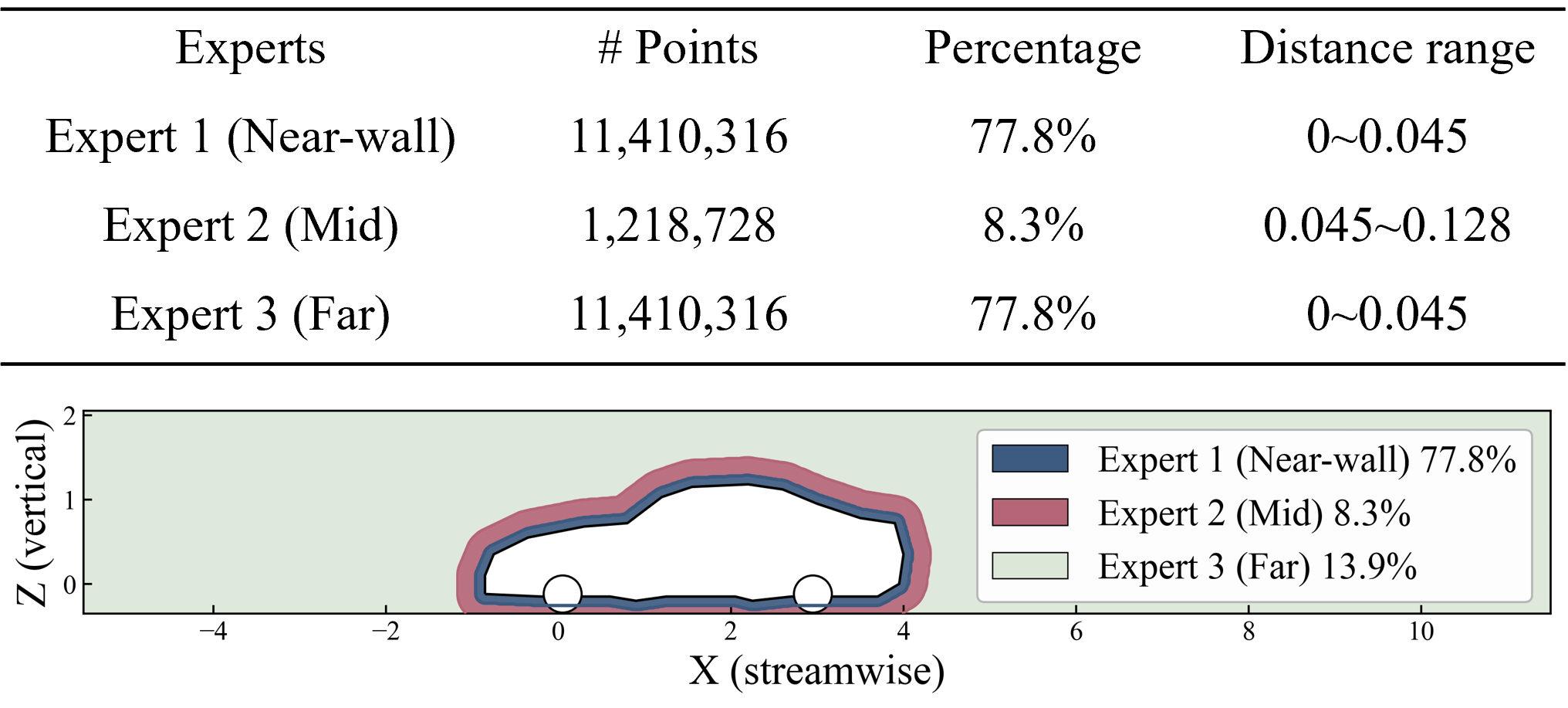}
        \caption{Dominant volume expert at each spatial location on
        a streamwise slice of the validation domain. The three
        experts partition the volume into a near-wall band
        (expert~1, 77.8\%), a transition layer (expert~2, 8.3\%),
        and a free-stream region (expert~3, 13.9\%).}
        \label{fig:svmoe-emergence}
    \end{subfigure}
    \hfill
    \begin{subfigure}[t]{0.49\linewidth}
        \centering
        \includegraphics[width=\linewidth]{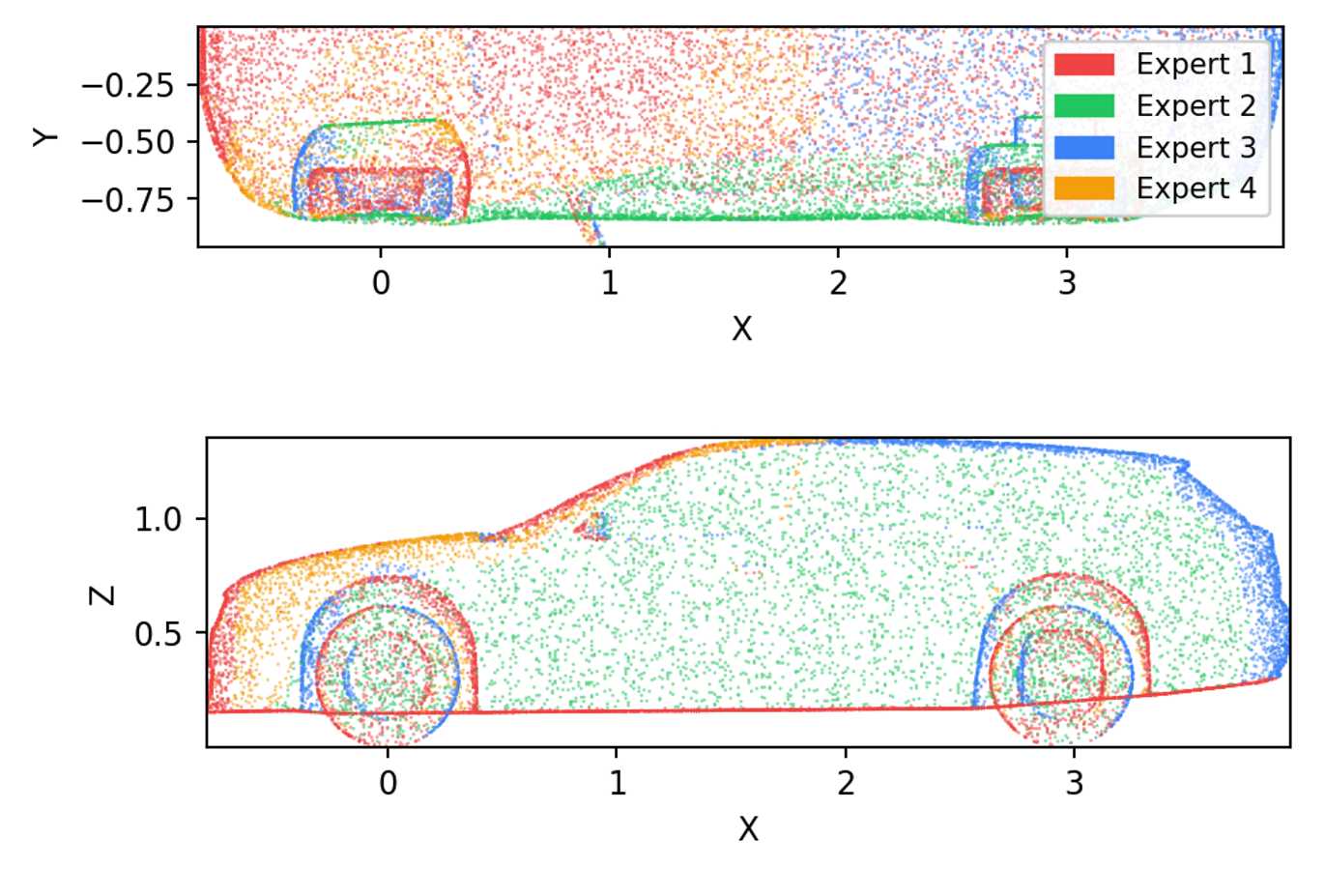}
        \caption{Dominant surface expert at each point on the
        vehicle body, shown from two views and colored by expert
        index.}
        \label{fig:svmoe-surface}
    \end{subfigure}\\[0.7em]
    \begin{subfigure}[t]{0.62\linewidth}
        \centering
        \includegraphics[width=\linewidth]{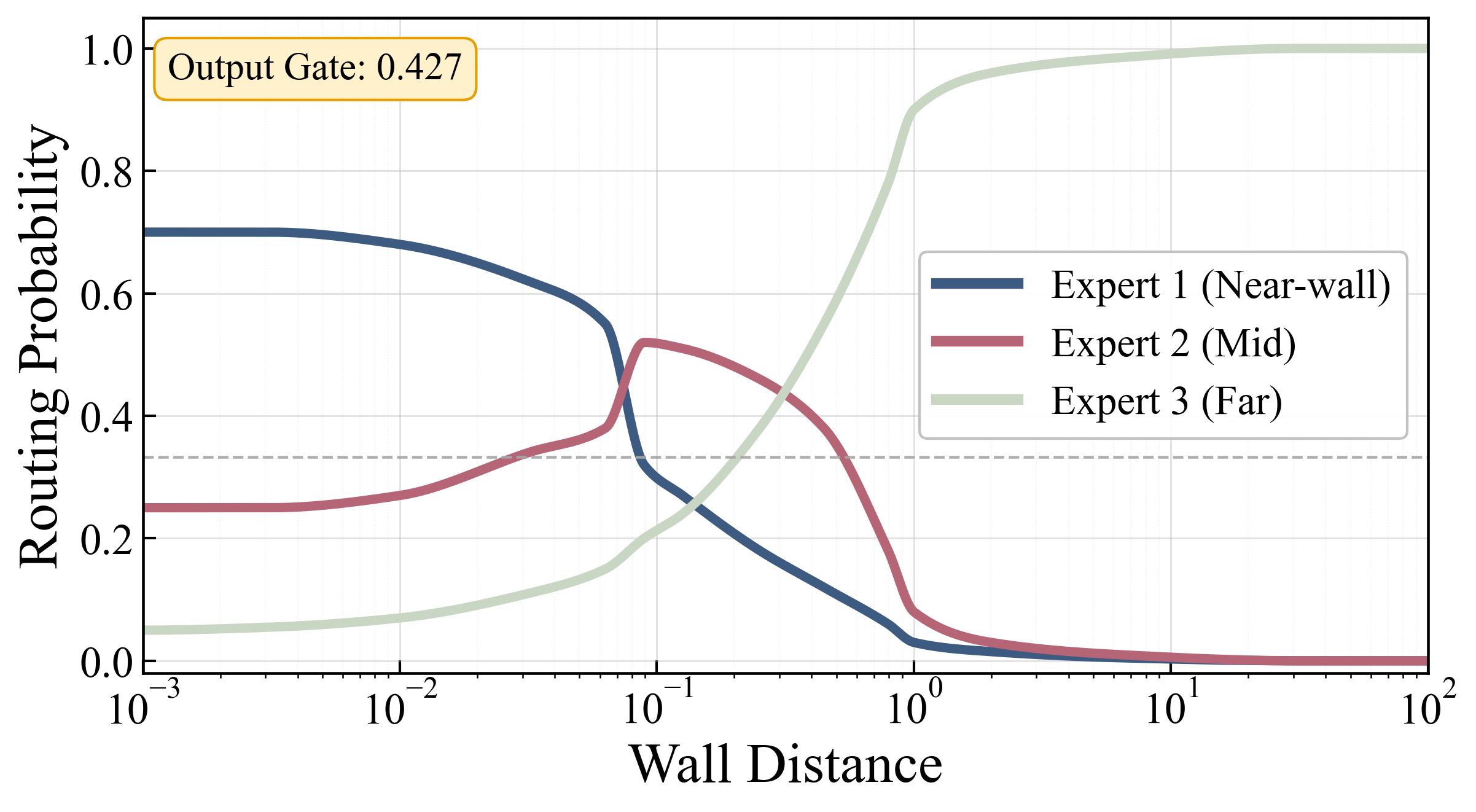}
        \caption{Routing distribution over $\dw$ during training:
        approximately uniform at epoch~1, progressively separating
        into near-wall, transition, and free-stream regions, and
        converging by epoch~1000.}
        \label{fig:svmoe-routing-evolution}
    \end{subfigure}
    \caption{Emergent expert structure on DrivAerML. The dominant
    volume expert at each spatial location
    (\subref{fig:svmoe-emergence}), the dominant surface expert on
    the body (\subref{fig:svmoe-surface}), and the evolution of the
    volume routing distribution over $\dw$ from epoch~1 to
    epoch~1000 (\subref{fig:svmoe-routing-evolution}). The routers'
    only inputs are $\dw$ and $\phi_s$, and no supervision is
    provided on the assignment.}
    \label{fig:svmoe-volume-analysis}
\end{figure}

Aggregated over the validation set, the three volume experts
dominate disjoint intervals of $\dw$
(Figure~\ref{fig:svmoe-volume-analysis}\subref{fig:svmoe-emergence}).
Expert~1 dominates the near-wall band ($\dw < 0.045$) and carries
77.8\% of the routing mass, the fraction of volume tokens for which
this expert receives the top gate weight; expert~2 covers the
intermediate transition region ($0.045 \le \dw \le 0.128$) and
carries 8.3\%; and expert~3 dominates the free stream
($\dw > 0.128$) and carries 13.9\%. The partition matches the
expected three-zone structure of wall-bounded turbulence, with a
thin but data-dense near-wall region, a small transition band, and
an outer region. The surface branch
organizes in the same way: the dominant surface expert at each
point on the vehicle body
(Figure~\ref{fig:svmoe-volume-analysis}\subref{fig:svmoe-surface})
groups
geometrically similar regions under the same expert, partitioning
the body into coherent geometric regions; the surface mixture is
active in all vehicle-aerodynamics experiments. The volume
partition is, moreover, learned rather than imposed by
initialization:
Figure~\ref{fig:svmoe-volume-analysis}\subref{fig:svmoe-routing-evolution}
tracks the routing distribution over $\dw$ during training, and at
epoch~1 it is approximately uniform across all three experts at
every value of $\dw$, consistent with the small random
initialization of the router MLP. As training proceeds the
distribution concentrates gradually, a near-wall preference
emerging first and the three regions then separating, until by
epoch~1000 the routing has converged, with boundary locations
consistent with the canonical boundary-layer stratification rather
than with arbitrary thresholds in input space.

Because the routers see only $\dw$ and $\phi_s$ and no
expert-assignment target is provided, the result supports the
interpretation that the prescribed physical variables serve as the
axes along which the partitions are learned. Two qualifications
bound this reading: we use the term emergence in a narrow sense,
the organization of a single prescribed axis into contiguous
regions rather than the discovery of higher-dimensional structure,
and both a quantitative comparison of the learned volume boundaries
($\dw = 0.045$ and $0.128$) against the canonical $y^+$ thresholds
of boundary-layer theory and a detailed analysis of the
surface-expert structure are left to future work.

\begin{figure}[!htbp]
    \centering
    \includegraphics[width=0.95\linewidth]{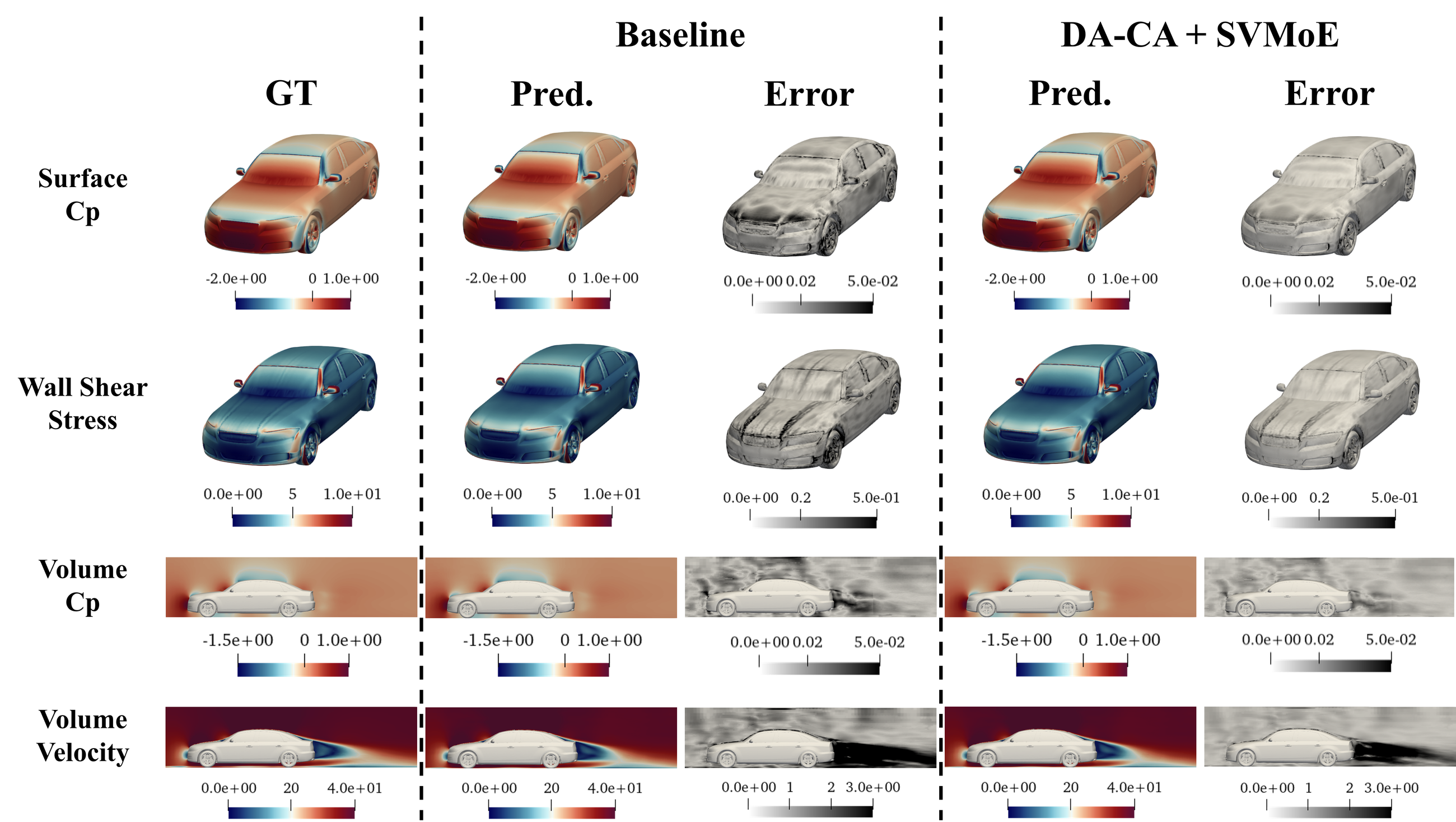}
    \caption{Qualitative field comparison between the unconditioned
    baseline, that is AB-UPT~\citep{Alkin2025ABUPT} without either
    proposed component, and the proposed model on a DrivAerML
    validation sample.}
    \label{fig:svmoe-fields}
\end{figure}

The expert structure is accompanied by lower prediction error.
Figure~\ref{fig:svmoe-fields} compares the
unconditioned baseline and the proposed model on a held-out validation sample
from DrivAerML. The pressure coefficient $\Cp$ and wall shear
stress $\WSS$ show lower surface errors with the proposed model, particularly
along the near-wall band. The volume velocity slice also improves
in the rear-wake region that was less affected by DA-CA: the wake
structure is sharper and the error band along the wake centerline is
narrower under the proposed model than under the baseline. The wake
lies in the outer band, so it is reached by the expert serving that
band rather than by one dedicated to it. Quantitatively,
the proposed model reduces the relative $L_2$ error on all four field
quantities relative to the baseline, with the largest improvement
on surface pressure and wall shear stress
(Table~\ref{tab:svmoe-ablation}, top). The wall-distance
decomposition of the velocity error
(Table~\ref{tab:svmoe-ablation}, bottom) makes the division of
labor explicit: DA-CA improves the near-wall zone from 12.5\% to
11.3\% but regresses in the far zone from 5.5\% to 6.1\%,
consistent with the wake gap identified in
Section~\ref{sec:daca-results}, whereas adding SVMoE recovers the
far zone to 5.3\%, below the baseline, while lowering the
near-wall error further to 10.9\%. The two components are therefore
complementary: DA-CA improves
spatial information aggregation in wall-proximal regions, while
SVMoE gives the outer band its own nonlinear transformation, which
benefits the far field and the wake within it.

\begin{table}[!htbp]
\centering
\caption{Cumulative ablation on DrivAerML (relative $L_2$) under the
50-sample regime, averaged over the validation set. Top: errors on
the four field quantities. Bottom: volume-velocity error decomposed
by wall-distance zone, with each zone's share of evaluation points.
Parentheses give the relative change with respect to the baseline.}
\label{tab:svmoe-ablation}
\footnotesize
\setlength{\tabcolsep}{4pt}
\begin{tabularx}{\linewidth}{@{}lYYYY@{}}
\toprule
Model & Surface $\Cp$ & WSS & Volume $\Cp$ & Velocity \\
\midrule
AB-UPT                      & 6.89\% & 11.40\% & 5.94\% & 8.23\% \\
DA-CA                       & 6.46\% ($-$6.2\%) & 10.63\% ($-$6.8\%) & 5.34\% ($-$10.1\%) & 7.73\% ($-$6.1\%) \\
\textbf{DA-CA + SVMoE} & \textbf{6.14\%} \textbf{($-$10.9\%)} & \textbf{10.23\%} \textbf{($-$10.3\%)} & \textbf{5.20\%} \textbf{($-$12.5\%)} & \textbf{7.40\%} \textbf{($-$10.1\%)} \\
\bottomrule
\end{tabularx}

\medskip

\begin{tabularx}{\linewidth}{@{}lYYYY@{}}
\toprule
Zone (Velocity) & Point share & AB-UPT & DA-CA & DA-CA + SVMoE \\
\midrule
Near-wall & 61\% & 12.5\% & 11.3\% ($-$9.6\%) & 10.9\% ($-$12.8\%) \\
Mid       & 23\% & 6.3\%  & 6.0\% ($-$4.8\%)  & 5.8\% ($-$7.9\%) \\
Far       & 16\% & 5.5\%  & 6.1\% ($+$10.9\%) & 5.3\% ($-$3.6\%) \\
\midrule
\textbf{Overall} & \textbf{100\%} & \textbf{8.23\%} & \textbf{7.73\% ($-$6.1\%)} & \textbf{7.40\% ($-$10.1\%)} \\
\bottomrule
\end{tabularx}
\end{table}

The improvement is not specific to the 50-sample budget, nor to the
backbone on which it was developed. To probe the data-rich end of the
practical range, and at the same time to test whether the two
mechanisms depend on the anchor construction of AB-UPT, we retrained
four models on 300 DrivAerML training samples with 50 held-out test
cases, keeping all other hyperparameters fixed: each of the two
backbones, AB-UPT and Transolver-3, with and without the proposed
components (Table~\ref{tab:data300}). On AB-UPT the full model
improves every field quantity, most strongly on volume pressure,
which drops from 5.95\% to 3.98\% ($-$33.1\%); the volume velocity
error falls from 8.07\% to 6.57\% ($-$18.6\%) and the surface
quantities improve by 20.6\% and
15.2\%. Applied to Transolver-3 without any retuning, the same
components reduce all four errors by a similar and more uniform
margin, by up to 22.6\%, with the volume pressure error
falling from 6.81\% to 5.35\% ($-$21.4\%) and the volume velocity
error from 5.91\% to 4.65\% ($-$21.3\%). The two backbones do not
rank the same way on every quantity: Transolver-3 is the stronger of
the two on volume velocity and the weaker on volume pressure, and the
ordering of the conditioned models follows that of their backbones.
Within each backbone, however, the conditioning improves every field
quantity. It therefore remains effective at this larger training
scale and on both backbones, contributing on top of rather than in
place of additional data, and the direction of the effect does not
depend on whether the compressed representation is a set of anchor
tokens or a set of physical-state slices.

\begin{table}[!htbp]
\centering
\caption{DrivAerML test errors (relative $L_2$) under an enlarged
training set of 300 samples with 50 held-out test cases, for both
backbones with and without the proposed components, with
hyperparameters matched to the 50-sample protocol of
Table~\ref{tab:svmoe-ablation}. The held-out set differs from the
one used in Table~\ref{tab:svmoe-ablation}, so absolute errors are
not directly comparable across the two tables; the comparison of
interest is between each backbone and its conditioned counterpart
within this table, which are trained and evaluated on the same split.
Here ``Ours'' denotes DA-CA together with SVMoE, the same pair
ablated in Table~\ref{tab:svmoe-ablation}. Parentheses give the
relative reduction in error with respect to the corresponding
unmodified backbone.}
\label{tab:data300}
\footnotesize
\setlength{\tabcolsep}{4pt}
\begin{tabularx}{\linewidth}{@{}lYYYY@{}}
\toprule
Model & Surface $\Cp$ & WSS & Volume $\Cp$ & Velocity \\
\midrule
AB-UPT & 5.19\% & 9.43\% & 5.95\% & 8.07\% \\
\textbf{AB-UPT + Ours} & \textbf{4.12\%} \textbf{($-$20.6\%)} & \textbf{8.00\%} \textbf{($-$15.2\%)} & \textbf{3.98\%} \textbf{($-$33.1\%)} & \textbf{6.57\%} \textbf{($-$18.6\%)} \\
\midrule
Transolver-3 & 5.48\% & 9.84\% & 6.81\% & 5.91\% \\
\textbf{Transolver-3 + Ours} & \textbf{4.24\%} \textbf{($-$22.6\%)} & \textbf{7.85\%} \textbf{($-$20.2\%)} & \textbf{5.35\%} \textbf{($-$21.4\%)} & \textbf{4.65\%} \textbf{($-$21.3\%)} \\
\bottomrule
\end{tabularx}
\end{table}

The improvement is also not specific to a training set that covers
the design space uniformly. The training and test cases used above
are drawn at random, so both cover the same part of that space. As a
harder check, we ranked all 473 available runs by their first
principal component in the 16-dimensional space of standardized
geometric design parameters, trained on the 50 designs at one end and
tested on the 50 at the other. The two ranges do not overlap and are
separated by 2.63 population standard deviations of the score, so the
test geometries fall outside the range seen during training, and the
ranking uses the geometry alone, with no reference to the flow fields
or to either model's error. The baseline was retrained on the same
split under the same budget and hyperparameters. Errors roughly
double for both models relative to the random split
(Table~\ref{tab:ood-pc1}), and the proposed model improves on all
four field quantities, by up to 26.2\%, about twice its gain in
distribution, with the lower error on all 50 test designs. One
caveat applies: the split narrows the training
distribution as well as displacing it, since the 50 designs at one
end resemble one another more closely than a random 50 would, so part
of the enlarged margin may reflect the reduced diversity of the
training set rather than the displacement alone.

\begin{table}[!htbp]
\centering
\caption{DrivAerML design-space extrapolation split (relative $L_2$).
Training uses the 50 designs at one end of the first principal
component of the standardized geometric design parameters and testing
the 50 at the other, with the training budget and all hyperparameters
matched to the random-split protocol of
Table~\ref{tab:svmoe-ablation}. Parentheses give the relative
reduction in error with respect to the baseline. The
proposed model attains the lower error on all 50 test designs
(two-sided sign test, $p = 1.8 \times 10^{-15}$).}
\label{tab:ood-pc1}
\footnotesize
\setlength{\tabcolsep}{4pt}
\begin{tabularx}{\linewidth}{@{}lYYYY@{}}
\toprule
Model & Surface $\Cp$ & WSS & Volume $\Cp$ & Velocity \\
\midrule
AB-UPT & 15.64\% & 24.63\% & 13.64\% & 18.28\% \\
\textbf{DA-CA + SVMoE} & \textbf{11.54\%} \textbf{($-$26.2\%)} & \textbf{18.92\%} \textbf{($-$23.2\%)} & \textbf{10.56\%} \textbf{($-$22.6\%)} & \textbf{14.56\%} \textbf{($-$20.4\%)} \\
\bottomrule
\end{tabularx}
\end{table}

Finally, we compare the proposed model with representative 3D
flow-field surrogates under the identical 50-sample DrivAerML
protocol, with every method trained on the same train/test split and
the same number of sampled points per case. Training time is not
equalized; under this protocol Transolver required more than 1.5
times the wall-clock training time of AB-UPT, so the ranking is not
driven by a larger training budget for the proposed model. As shown
in Table~\ref{tab:sota}, the proposed model attains the best accuracy
on every field quantity in this low-data regime. Two factors
contribute, and we keep them separate: the AB-UPT backbone is already
competitive with the other surrogates at 50 samples, so part of the
margin reflects the backbone rather than our interventions, while the
contribution of the components themselves is the further reduction of
up to 12.5\% isolated by the ablation in
Table~\ref{tab:svmoe-ablation}, and of up to 26.2\% under the
extrapolation split of Table~\ref{tab:ood-pc1}.

\begin{table}[!htb]
\centering
\caption{Comparison against prior 3D flow-field surrogates on
DrivAerML (relative $L_2$), under the identical 50-sample protocol
with the same train/test split and the same number of sampled
training points per case. The proposed model attains the
best accuracy on every field quantity in this low-data regime. For
each quantity, the lowest error among the prior surrogates is set
in bold, and parentheses in the last row give the relative
reduction of the proposed model with respect to that best prior
result.
}
\label{tab:sota}
\footnotesize
\setlength{\tabcolsep}{4pt}
\begin{tabularx}{\linewidth}{@{}lYYYY@{}}
\toprule
Model & Surface $\Cp$ & WSS & Volume $\Cp$ & Velocity \\
\midrule
AB-UPT & 6.89\% & 11.40\% & \textbf{5.94\%} & 8.23\% \\
UPT                            & 10.63\% & 17.13\% & 7.68\% & 11.34\% \\
Transolver                     & 9.58\% & 15.21\% & 8.59\% & 12.41\% \\
Transolver-3                   & 6.90\% & \textbf{11.02\%} & 5.98\% & 7.81\% \\
GeoTransolver     & \textbf{6.67\%} & 15.44\% & 6.03\% & \textbf{8.18\%} \\
\midrule
\textbf{Ours} & \textbf{6.14\%} \textbf{($-$7.9\%)} & \textbf{10.23\%} \textbf{($-$7.2\%)} & \textbf{5.20\%} \textbf{($-$12.5\%)} & \textbf{7.40\%} \textbf{($-$9.5\%)} \\
\bottomrule
\end{tabularx}
\end{table}

\section{Generalization across Geometric Families}
\label{sec:ood}

DrivAerML, on which DA-CA and SVMoE are developed, contains a
single family of vehicle silhouettes around a common
parameterization. A reasonable concern is that the gains reported
in Sections~\ref{sec:daca} and~\ref{sec:svmoe} reflect features
of that family rather than mechanisms that transfer to new
geometries. The extrapolation split of
Section~\ref{sec:svmoe-results} displaced the design parameters while
holding the geometric family fixed; this section changes the family
itself, on DrivAerNet++. The proposed model is DA-CA + SVMoE as in
Section~\ref{sec:svmoe-results}.

\subsection{Geometric Out-of-Distribution Evaluation on DrivAerNet++}
\label{sec:ood-drivaernet}

\begin{figure}[!htbp]
    \centering
    \includegraphics[width=\linewidth]{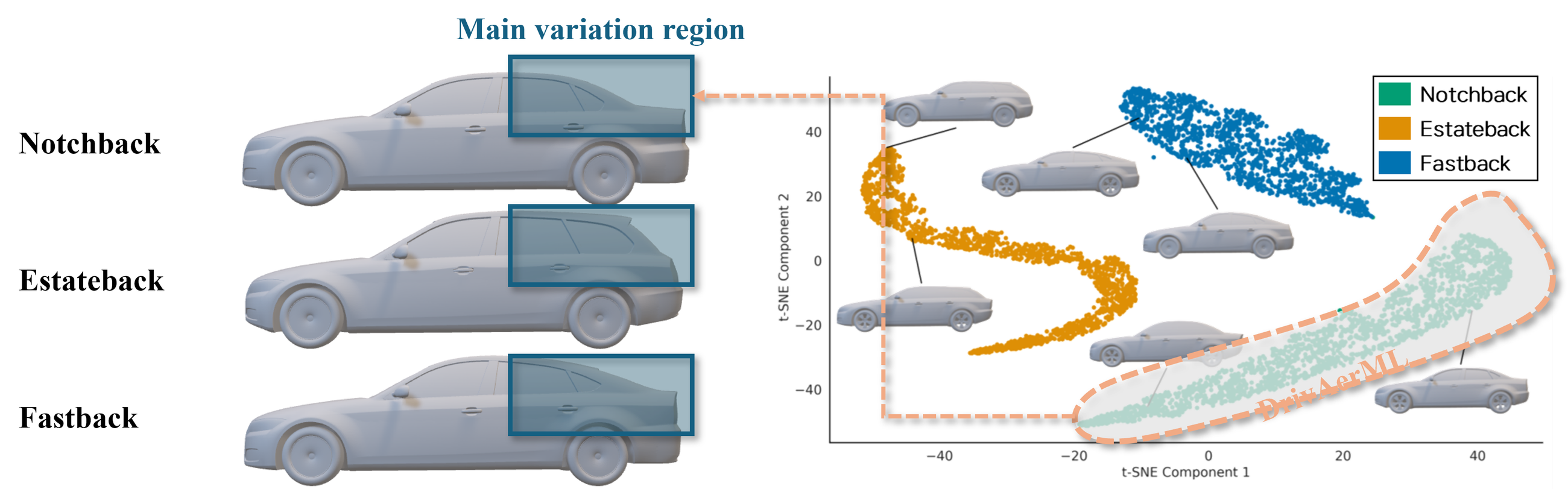}
    \caption{Geometric distribution shift between DrivAerML and
    DrivAerNet++.
    Left: rear-silhouette variations across the three DrivAerNet++
    body types, with the principal variation localized near the
    rear roof. Right: t-SNE embedding of a shared geometric
    descriptor over all samples. DrivAerML forms one cluster,
    DrivAerNet++ forms three clusters separated by body type, and
    the two benchmarks overlap only in a small region of the
    embedding.}
    \label{fig:ood-tsne}
\end{figure}

Figure~\ref{fig:ood-tsne} visualizes the geometric distribution
shift between the two benchmarks via a t-SNE embedding of a shared
geometric descriptor. DrivAerML forms a single narrow cluster,
consistent with its morphing-based construction around a common
silhouette, while DrivAerNet++ forms three separated clusters
corresponding to the Fastback, Notchback, and Estateback body types,
overlapping DrivAerML in only one region. The principal variation
across the three types lies in the rear quarter, where the silhouette
transitions from a coupe-like profile to a sedan rear deck to an
elongated rear roofline. It is therefore local to a small fraction of
the surface, but it changes the wake topology and the rear-pressure
recovery, both of which the surrogate must predict. The setting is a
clean test of whether the inductive biases transfer across geometric
families without retraining the routing structure.

\subsubsection{Protocol}
\label{sec:ood-protocol}

The DrivAerNet++ subset used in this work contains 688 Estateback,
692 Fastback, and 386 Notchback samples. We evaluate the two
complementary protocols summarized in
Table~\ref{tab:ood-protocol}. Case~1 pools all three body types and
splits each type into training and test sets. Because every body type
appears in training, this case asks whether the gains survive a larger
and more heterogeneous training set, not whether they survive an
unseen shape. Cases~2A--2C (Leave-One-Body-Out, LOBO) hold out one of
the three body types entirely: they train on 25 samples each of the
two remaining types and evaluate on the third, so the test silhouette
is one the model has never seen. Their
50-sample budget matches the DrivAerML protocol of
Section~\ref{sec:daca}, so any change in error relative to
Section~\ref{sec:svmoe-results} reflects the unseen geometry rather
than a change in the amount of training data. Under both protocols we
compare the proposed model, DA-CA + SVMoE, to the unconditioned
baseline under matched hyperparameters and matched compute budget,
and report the same four field quantities used on DrivAerML in
relative $L_2$ error.

\begin{table}[!htbp]
\centering
\caption{DrivAerNet++ evaluation protocols. Case~1 pools all three
body types with a per-type train--test split; Cases~2A--2C
each hold out one body type (Leave-One-Body-Out) and train on
25 samples of each remaining type, matching the 50-sample budget
of Section~\ref{sec:daca}.}
\label{tab:ood-protocol}
\footnotesize
\setlength{\tabcolsep}{4pt}
\begin{tabularx}{\linewidth}{@{}lYYcc@{}}
\toprule
& \multicolumn{2}{c}{Body types} & \multicolumn{2}{c}{Samples} \\
\cmidrule(lr){2-3} \cmidrule(l){4-5}
Case & Train & Test & Train & Test \\
\midrule
1 (multi-body) & Estateback (550), Fastback (554),\newline Notchback (309)
& Estateback (138), Fastback (138),\newline Notchback (77) & 1413 & 353 \\
2A (LOBO) & Fastback (25), Notchback (25) & Estateback (25) & 50 & 25 \\
2B (LOBO) & Estateback (25), Notchback (25) & Fastback (25) & 50 & 25 \\
2C (LOBO) & Estateback (25), Fastback (25) & Notchback (25) & 50 & 25 \\
\bottomrule
\end{tabularx}
\end{table}

\subsubsection{Case 1: Multi-Body Results}
\label{sec:ood-case1}

\begin{figure}[!htbp]
    \centering
    \includegraphics[width=0.85\linewidth]{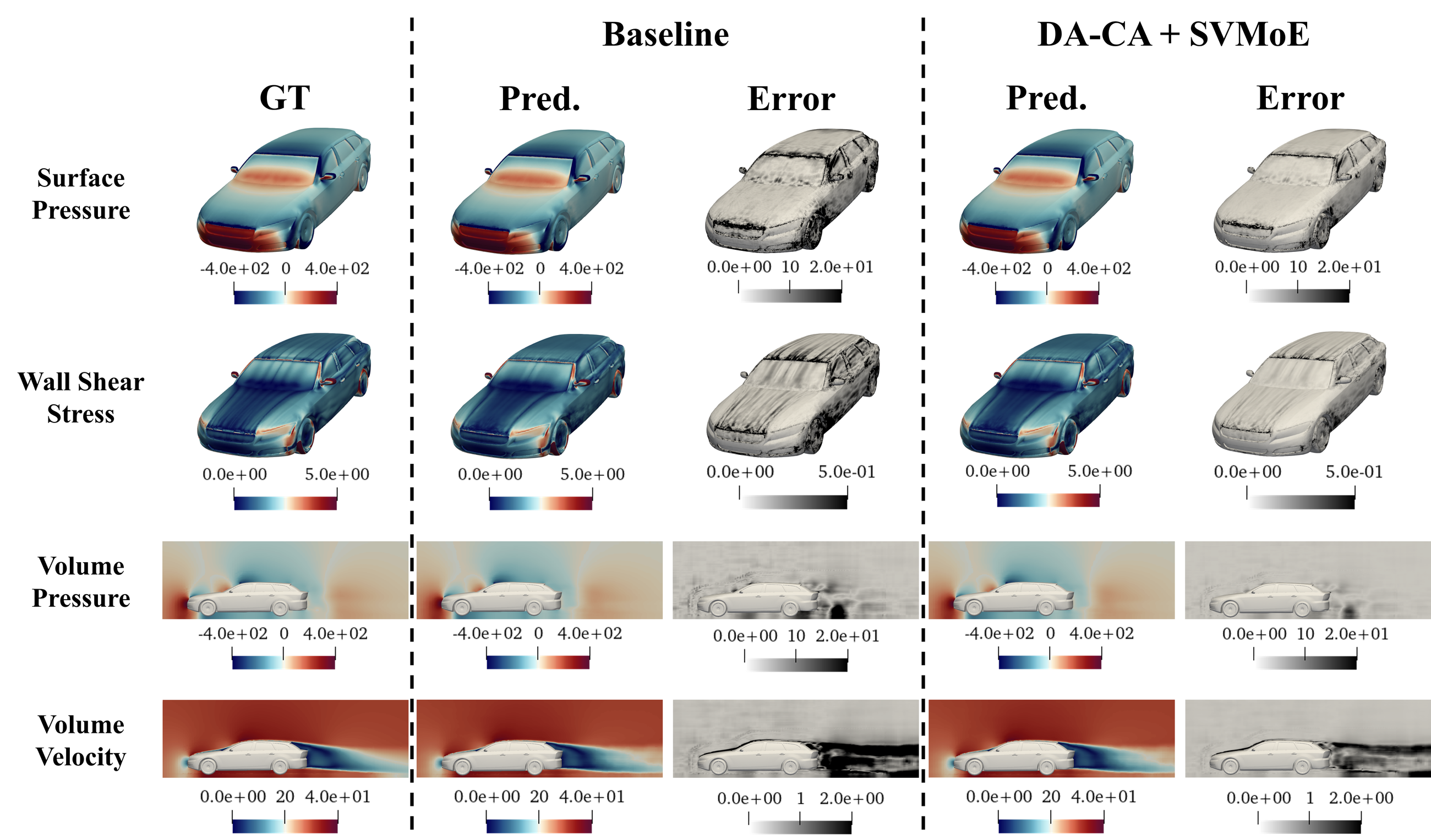}
    \caption{Case~1 field comparison on DrivAerNet++. Predicted
    fields and errors are shown for a test sample of each
    body type, with the unconditioned baseline and the proposed
    model (DA-CA + SVMoE) placed side by side. The proposed model
    consistently reduces surface pressure and
    wall shear stress error along the body and produces more
    localized volume-velocity error patterns.}
    \label{fig:ood-case1-fields}
\end{figure}

Under Case~1, where the training set spans all three body types,
the proposed model improves on the unconditioned baseline across all
four field quantities and all three body silhouettes.
Figure~\ref{fig:ood-case1-fields} shows the qualitative comparison
for a representative test sample of each body type, and
Table~\ref{tab:ood-case1} reports the corresponding errors. The
absolute errors are higher than in the DrivAerML regime of
Section~\ref{sec:daca}, reflecting the geometric heterogeneity of the
pooled three-body training set rather than a change in the ordering
of the two models.

\begin{table}[!htbp]
\centering
\caption{DrivAerNet++ Case~1 (multi-body) test errors
(relative $L_2$). Training and test sets both span all
three body types. The proposed model (DA-CA + SVMoE) improves on the
unconditioned baseline across all four field quantities. Parentheses
give the relative reduction in error.}
\label{tab:ood-case1}
\footnotesize
\setlength{\tabcolsep}{4pt}
\begin{tabularx}{\linewidth}{@{}lYYYY@{}}
\toprule
Model & \shortstack{Surface\\Pressure} & WSS & \shortstack{Volume\\Pressure} & \shortstack{Volume\\Velocity} \\
\midrule
AB-UPT        & 15.69\% & 23.21\% & 15.33\% & 12.64\% \\
\textbf{DA-CA + SVMoE} & \textbf{13.60\%} \textbf{($-$13.3\%)} & \textbf{21.54\%} \textbf{($-$7.2\%)} & \textbf{13.43\%} \textbf{($-$12.4\%)} & \textbf{10.55\%} \textbf{($-$16.5\%)} \\
\bottomrule
\end{tabularx}
\end{table}

\subsubsection{Case 2: Leave-One-Body-Out Results}
\label{sec:ood-case2}

\begin{figure}[!htbp]
    \centering
    \includegraphics[width=\linewidth]{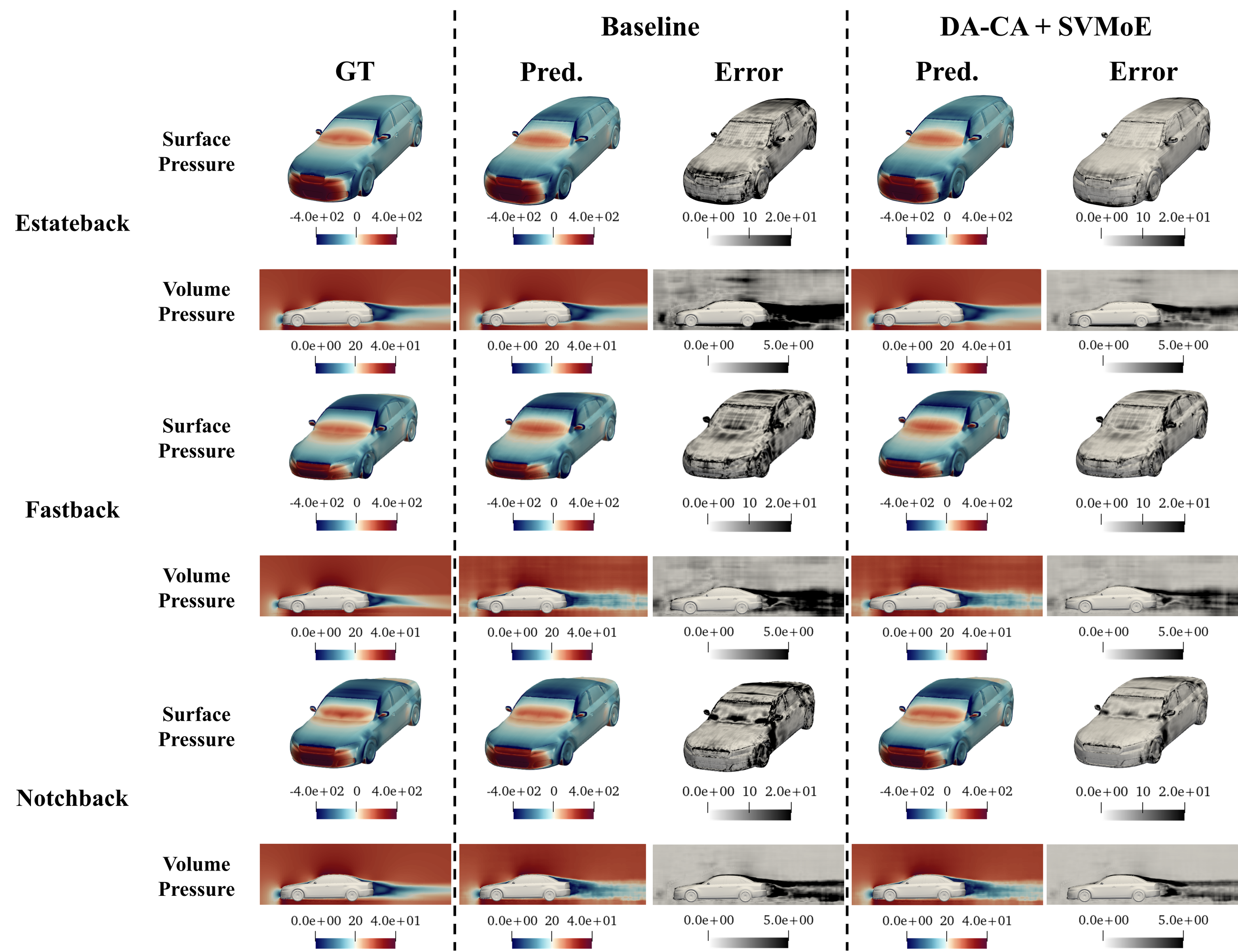}
    \caption{Leave-One-Body-Out field comparison on
    DrivAerNet++. Each row corresponds to one held-out body type
    (Estateback, Fastback, Notchback) and shows the predicted
    fields and errors for the unconditioned baseline and the
    proposed model (DA-CA + SVMoE), along with the per-quantity
    improvement. The relative ordering
    of the two models is preserved across all three held-out body
    types.}
    \label{fig:ood-case2-fields}
\end{figure}

Case~2 is the more stringent test. Each sub-case removes an entire
body type from training, so the model must predict on a geometric
silhouette it has never seen.
Figure~\ref{fig:ood-case2-fields} shows the field comparison for
each held-out body type, and Table~\ref{tab:ood-lobo} reports the
test errors for the three sub-cases. The proposed model improves on
the baseline in all twelve cells, by up to 14.2\%, so the gain is
localized neither to a particular body type nor to a particular
quantity. The error magnitudes are substantially higher than in the
DrivAerML regime of Section~\ref{sec:daca}, as expected for genuinely
out-of-distribution evaluation under a 50-sample budget, but the
ordering of the two models is preserved throughout.

\begin{table}[!htbp]
\centering
\caption{DrivAerNet++ Leave-One-Body-Out test errors
(relative $L_2$). Each row holds out one body type from training
and reports prediction error on that type. The proposed model
(DA-CA + SVMoE) improves on the unconditioned baseline in every cell. Parentheses give the relative
reduction in error.}
\label{tab:ood-lobo}
\footnotesize
\setlength{\tabcolsep}{4pt}
\begin{tabularx}{\linewidth}{@{}lccYYYY@{}}
\toprule
Case & Held-out & Model & \shortstack{Surface\\Pressure} & WSS & \shortstack{Volume\\Pressure} & \shortstack{Volume\\Velocity} \\
\midrule
\multirow{2}{*}{2A} & \multirow{2}{*}{Estateback} & AB-UPT        & 22.32\% & 28.42\% & 22.83\% & 19.60\% \\
 & & \textbf{DA-CA + SVMoE} & \textbf{20.19\%} \textbf{($-$9.5\%)} & \textbf{27.03\%} \textbf{($-$4.9\%)} & \textbf{21.46\%} \textbf{($-$6.0\%)} & \textbf{18.05\%} \textbf{($-$7.9\%)} \\
\midrule
\multirow{2}{*}{2B} & \multirow{2}{*}{Fastback} & AB-UPT        & 17.59\% & 27.03\% & 19.04\% & 14.82\% \\
 & & \textbf{DA-CA + SVMoE} & \textbf{16.73\%} \textbf{($-$4.9\%)} & \textbf{26.35\%} \textbf{($-$2.5\%)} & \textbf{16.52\%} \textbf{($-$13.2\%)} & \textbf{13.31\%} \textbf{($-$10.2\%)} \\
\midrule
\multirow{2}{*}{2C} & \multirow{2}{*}{Notchback} & AB-UPT        & 20.35\% & 29.15\% & 19.45\% & 14.71\% \\
 & & \textbf{DA-CA + SVMoE} & \textbf{17.51\%} \textbf{($-$14.0\%)} & \textbf{27.64\%} \textbf{($-$5.2\%)} & \textbf{16.69\%} \textbf{($-$14.2\%)} & \textbf{13.45\%} \textbf{($-$8.6\%)} \\
\bottomrule
\end{tabularx}
\end{table}

\paragraph{Routing stability as a mechanism for OOD success.}
To ask whether the OOD improvement reflects a property of the routing
structure itself, we examine the expert routing distribution on the
held-out test sets. Across all three LOBO sub-cases, the volume
routing preserves the near-wall, transition, and free-stream
proportions that emerged on DrivAerML, each expert retaining its
share to within a few percent. The router sees $\dw$ alone and is
never exposed to body-type information, so this stability is
consistent with a wall-bounded-flow interpretation of the partition
rather than a silhouette-specific one.

\FloatBarrier

\section{Conclusion}
\label{sec:conclusion}

Transformer surrogates for 3D flow prediction take geometry as an
input, but they do not use it to decide how each point is processed.
Every point retrieves information from the compressed representation
by learned feature similarity alone, and every point is then passed
through the same feed-forward network. As a result, a point in the
boundary layer and a point in the free stream, or a point on the
surface and a point in the volume, are treated identically even
though the physics governing them differs. One might expect the
model to learn this distinction from data, but our baseline
experiments on DrivAerML show that it does not. Its cross-attention
weights are nearly flat in distance, so a point reads from nearby
and distant anchors alike; its error is concentrated in the
near-wall band, exactly where this indifference matters most; and
adding training data beyond roughly 50 cases does not change either
observation, which indicates that the cause is the architecture
rather than the amount of data. This paper therefore builds the
distinction into the architecture at the two points where it is
missing. Distance-Aware Cross-Attention (DA-CA) makes the retrieval
depend on the point's wall distance, so that near-wall and
outer-flow points gather different geometric information. This
improves the region where the baseline error is largest, at the cost
of a slight degradation in the far zone, because the retrieved
features are still processed by one transformation. Surface-Volume
Mixture-of-Experts (SVMoE) then makes the transformation depend on
the flow regime as well, routing volume points by wall distance and
surface points by local geometry, so that the far-zone accuracy
traded away by DA-CA is recovered without losing the near-wall
improvement.

The main quantitative findings are as follows.

\begin{enumerate}[leftmargin=1.6em,label=\arabic*.]
\item \textbf{DA-CA and SVMoE.} On DrivAerML with 50 training
samples, conditioning cross-attention on wall distance reduces the
volume pressure error by 10.1\%, but at the cost of the outer flow: the volume velocity error falls by 9.6\% in the near-wall
zone and rises by 10.9\% in the far zone. Routing the feed-forward
layer by wall distance removes that cost. The far-zone error moves to
3.6\% below the baseline, the near-wall zone improves further to
12.8\% below, and the volume pressure error reaches 12.5\% below.
Given no supervision on the assignment, the volume experts settle
into near-wall (77.8\%), transition (8.3\%), and free-stream (13.9\%)
bands.

\item \textbf{Data scale and backbone.} The results above are
obtained at the data budget the method targets and on the backbone
it was developed on. To check that the advantage is tied to neither,
we retrained both backbones, with and without the components, on 300
training samples under the identical protocol. The advantage neither
fades with more data nor depends on the backbone: the same components
reduce the volume pressure and velocity errors by 33.1\% and 18.6\%
on AB-UPT, and by 21.4\% and 21.3\% on Transolver-3.

\item \textbf{Generalization.} On the DrivAerNet++ dataset, the
model improves volume pressure prediction by up to 14.2\% on car
body types that were not used in training, and the expert routing
does not change on these unseen types.
\end{enumerate}

Three limitations should be noted. First, all experiments use
steady-state flows and training sets of at most 300 cases; the
300-sample experiment shows that the gains persist at moderate
scale, but whether they hold on substantially larger datasets or on
unsteady flows is not established here. Second, wall distance is a
single scalar. It organizes the boundary layer well, but a point in
the wake and a point in the free stream at the same distance from
the wall are treated alike by both mechanisms, so wake regions,
whose dynamics also depend on vorticity and turbulent transport,
are not specifically addressed. Third, the out-of-distribution
evaluation of Section~\ref{sec:ood} was run on the
AB-UPT instantiation only; the evidence that the mechanisms are not
tied to one backbone comes from the in-distribution comparison of
Table~\ref{tab:data300}.

\section*{Acknowledgements}

This research was supported by the National Research Council of
Science \& Technology (NST) grant by the Korea government (MSIT)
(No.~GTL24033-000).


\end{document}